\documentclass[a4paper,fleqn]{cas-sc}

\usepackage[authoryear,longnamesfirst]{natbib}
\usepackage{url}
\usepackage{placeins}

\makeatletter
\ExplSyntaxOn
\cs_set:Npn \__first_footerline: { }
\ExplSyntaxOff
\makeatother

\def\tsc#1{\csdef{#1}{\textsc{\lowercase{#1}}\xspace}}
\tsc{WGM}
\tsc{QE}
\tsc{EP}
\tsc{PMS}
\tsc{BEC}
\tsc{DE}

\begin{document}
\let\WriteBookmarks\relax
\def\floatpagepagefraction{1}
\def\textpagefraction{.001}
\shorttitle{Do Vision Models See Like the Brain?}
\shortauthors{S. Baghel et~al.}

\title[mode=title]{Do Vision Model See Like the Brain? A Comparison Across EEG Encoding Model}

\nonumnote{This research is funded by Indian Institute of Technology Mandi,
Department of Higher Education, Ministry of Education, Government of India.}

\author[1]{Shashank Baghel}[type=editor,
                        auid=000,bioid=1,
                        prefix=,
                        role=,
                        orcid=0000-0002-5099-9524]

\ead{s25038@students.iitmandi.ac.in}

\credit{Conceptualization and Methodology, Experiments, Analysis, Draft preparation, Review and Editing}

\affiliation[1]{organization={Centre for Human-Computer
Interaction, Indian Institute of Technology},
                addressline={Mandi},
                city={Mandi},
                postcode={175075},
                state={Himachal Pradesh},
                country={India}}

\author[2,3]{Kshitij Dwivedi}[orcid=0000-0001-6442-7140]
\ead{kshitijdwivedi@google.com}

\credit{Conceptualization and Methodology, Review and Editing}

\author[1]{Dinesh Singh}[%
   role=,
   suffix=,
  orcid=0000-0001-8889-9847]
\cormark[1]
\ead{dineshsingh@iitmandi.ac.in}

\credit{Conceptualization and Methodology, Review and Editing}

\affiliation[2]{organization={Freie Universität},
                city={Berlin},
                country={Germany}}

\affiliation[3]{organization={Goethe University},
                city={Frankfurt},
                country={Germany}}

\author[1]{Sanjeev Nara}[orcid=0000-0002-9007-3160]
\cormark[1]
\ead{sanjeevnara@iitmandi.ac.in}

\credit{Conceptualization and Methodology, Review and Editing}

\cortext[cor1]{Equal Corresponding Authors}


\begin{abstract}
Convolutional neural networks (CNNs) and vision transformers are both used to model the human visual system, but whether the two architectures diverge at a specific point in network depth is unclear. We compared six CNNs and two vision transformers by computing the Pearson correlation ($r$) between each model's predicted and measured EEG response at every layer or block, in ten participants viewing 200 natural images. For the transformer models, we also tested four token representations, from the classification (CLS) token alone to CLS combined with all patch tokens. CNNs showed strongest correspondence at the earliest layers, weakening at deeper layers, particularly later in the post-stimulus response. Transformers instead sustained strong correspondence at their deepest blocks, though not at their earliest ones. This advantage depended on token representation: pooled representations gave weaker peak correlations ($r{\approx}0.48$--$0.51$) than representations retaining all patch tokens ($r{=}0.640$ for CLIP-ViT-B/32, $r{=}0.656$ for DINOv2-ViT-B/14). Controlled comparisons showed architecture, not training objective, drove this effect: MoCo-v1 and ResNet-50 (matched architecture) performed nearly identically ($r{=}0.673$, $0.670$), whereas CLIP-RN50 and CLIP-ViT-B/32 (matched objective) diverged until patch tokens were preserved. We propose that CNN training's classification bottleneck compresses brain-relevant information at depth, unlike transformers' self-attention and non-classification objectives. A spatial topography analysis showed a common occipital-dominant pattern across all models, indicating these differences reflect signal strength and persistence rather than distinct brain regions. Patch-preserving transformer representations sustain brain-predictive correspondence where CNNs collapse.
\end{abstract}


\maketitle

\section{Introduction}

Deep vision models have become the dominant tool for investigating how the brain constructs visual representations.
In the past years of work comparing convolutional neural networks (CNNs) to
neural recordings, a consistent picture has emerged that early network layers
predict early visual cortex, and deeper layers predict higher visual areas
\citep{yamins2014, khaligh2014, guclu2015, cichy2016}. This correspondence is
useful because it turns a question, whether a model processes images
the way the brain does, into a quantitative one that can be tested layer by
layer. The motivation runs in both directions, beyond using models to
explain the brain, the human visual system still outperforms vision models on many tasks such as few shot generalization and
robust multimodal understanding, so identifying where current models
diverge from biological vision is also a step toward designing more brain
like vision models. But nearly all of this evidence comes from CNNs, and nearly all of it
comes from fMRI or invasive electrophysiology. Vision transformers have
since become the dominant architecture in computer vision, trained under
objectives, classification, contrastive language image alignment, and self-distillation, that differ sharply from one another and from the
classification objective used in the CNN models. Whether the
brain correspondence found for CNNs is a property of the visual hierarchy
itself, or a property tied to CNNs and their classification objective
specifically, is the question this work answers. We ask it using EEG modality, does
the CNN to brain correspondence generalize to transformers, and if it does
not, does the divergence trace to architecture, to training objective, or
to both.

\begin{figure}[pos=htbp]
\centering
\includegraphics[width=\linewidth]{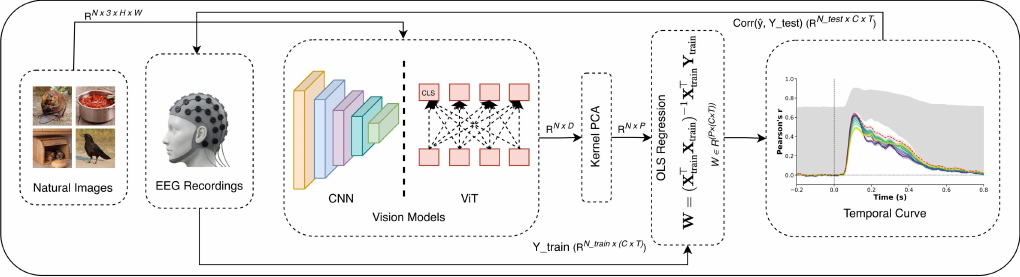}
\caption{Overview of the EEG encoding pipeline. Natural images are passed
through vision models (CNNs and ViTs) to extract features, which are
reduced via Kernel PCA and mapped to EEG responses via OLS regression.
Model predictions are evaluated against held-out EEG using Pearson
correlation.}
\label{fig:pipeline}
\end{figure}

The CNN side of this question is fairly explored. Supervised CNNs trained on ImageNet classification reliably
predict neural activity along the ventral stream, with a hierarchical
match between network depth and cortical stage \citep{yamins2014,
khaligh2014, guclu2015}. Brain Score and related benchmarking efforts have
extended this into a standard methodology for scoring any model against
primate visual cortex \citep{schrimpf2018}. MEG and EEG work has shown that this same hierarchical correspondence unfolds over time, with a feedforward sweep from low level to category level representations visible within the first 200 ms of the neural response \citep{cichy2016,
grootswagers2019}. Separately, a handful of recent studies have begun to
test vision transformers and multimodal models such as CLIP against fMRI,
generally finding that these models predict higher visual cortex at least
as well as, and sometimes better than, CNNs \citep{oota2022, wang2023}. What
is missing is the intersection of these two literatures: no prior study
evaluates a broad set of CNN and transformer models against EEG within a
single, matched analysis pipeline, and none uses the millisecond
resolution of EEG to ask specifically where, along network depth, CNNs and
transformers begin to disagree. We close this gap by performing a layer
by layer analysis of all models, evaluating Pearson's Correlation at every
individual layer or transformer block (Fig.~\ref{fig:pipeline}). This gives a natural way to ask when, not just whether, a
model diverges from the brain, using EEG's millisecond resolution to
resolve a layer by layer profile that fMRI's slow hemodynamic response
cannot. Applying this to six CNNs and two vision transformers reveals a
clear asymmetry. At the earliest and intermediate layers, the two
architecture families are statistically indistinguishable, predicting EEG
responses with almost similar peak Pearson correlation. The
divergence appears only once the network reaches its deeper layers: CNNs middle layers correlation with EEG begins to fall noticeably below that of the
transformers middle blocks, and the gap widens further at the final layers.
The result is a
gap that is negligible at shallow depth but grows into a clear, sizeable
transformer advantage by the final layers.

To separate why this happens, we test two candidate causes rather than
treating "transformer" as a single unexplained label. The first is the
classification bottleneck: a network trained to sort images into 1,000
ImageNet categories has every incentive to discard information that does
not help distinguish those categories, and this compression appears to
actively damage its correspondence with the brain's own semantic
representations \citep{yamins2014, khaligh2014}. CORnet-S offers a direct
test of whether this is fixable by architecture alone: its layers are
explicitly labelled V1, V2, V4, and IT to match cortical anatomy, yet its
IT layer still falls into the same low pearsons correlation range as the
deepest layers of every other CNN we tested. The second candidate cause is self attention itself:
every transformer block, including the last, attends across the full set
of spatial tokens, so information present early in the network remains
reachable throughout, unlike the CNN's progressive spatial pooling.
CLIP-ViT-B/32 is trained to align images with natural language
descriptions \citep{radford2021} and DINOv2 is trained with no labels at
all through self distillation \citep{oquab2023}; neither is forced to
compress its output into a fixed small set of categories. Because training
objective and self attention are conflated in a simple CNN versus
transformer comparison, we include two additional models, MoCo-v1 and
CLIP-RN50, that each hold one factor fixed while varying the other.

Contributions of this study includes:
\begin{itemize}
    \item \textbf{A layer by layer Analysis} We test six CNNs and two Vision Transformers across every
    individual layer (as shown in Fig.~\ref{fig:pipeline}) and show that the two architecture families predict
    EEG equally well at early and intermediate layers, then diverge
    sharply at deeper layers.
    \item \textbf{Two controlled experiments that isolate the cause of
    this divergence.} MoCo-v1 is compared against supervised CNNs of the
    same architecture to isolate the effect of training objective alone,
    and CLIP-RN50 is compared against CLIP-ViT-B/32 under the same
    training objective to isolate the effect of self-attention alone.
    \item \textbf{Exploring variation of Vision Transformers tokens } We argue that the CNN classification bottleneck and
    the transformer's token preserving self-attention together account
    for why transformers, but not CNNs, sustain brain like representations
    at the deepest layers, which correspond to inferotemporal cortex, the
    brain's semantic object recognition stage.
\end{itemize}

\section{Related Work}

The literature connecting deep vision models to brain activity divides
naturally along two axes: the neuroimaging modality used to record the
brain signals, and the model architecture being tested. We organize this
section along both. We first trace the biological inspiration behind
convolutional architectures and their growing divergence from human
perception, then cover CNN-brain correspondence as characterized through
spatial neuroimaging (fMRI and invasive electrophysiology) and through
temporal neuroimaging (MEG and EEG), before closing with the still thin
literature on vision transformers as brain models.

\subsection{Biological Inspiration and Modern Optimization Divergence}

The modern study of brain encoding is rooted in the pioneering work of \citet{hubel1962}, who demonstrated that neurons in the primary visual cortex (V1) respond selectively to simple visual features such as oriented edges and proposed a hierarchical organization of visual processing. Their discovery of simple and complex cells established the biological principle that increasingly complex visual representations emerge through successive cortical stages, an idea that later inspired hierarchical computational vision models and, ultimately, convolutional neural networks. Fukushima's Neocognitron \citep{fukushima1980} introduced alternating S-cell and C-cell layers for hierarchical feature extraction and shift-invariant recognition, providing the conceptual precursor to modern CNNs. \citet{serre2007} extended this biologically inspired framework through hierarchical feedforward models employing alternating template-matching and max-pooling operations for rapid object categorization, and \citet{krizhevsky2012}'s AlexNet demonstrated that these principles could scale to large-scale image classification. Subsequent work showed that convolutional networks develop increasingly specialized representations across their intermediate and deeper layers, with different layers contributing differently to classification performance \citep{zeiler2014}. However, networks optimized primarily for classification accuracy may diverge from human visual perception. Model-optimized controversial stimuli expose systematic differences between neural network predictions and human perceptual judgments \citep{golan2020}, while comprehensive reviews highlight persistent discrepancies between CNNs and biological vision \citep{lindsay2021}. Furthermore, recent work suggests that improvements in ImageNet accuracy do not necessarily correspond to better alignment with human perceptual similarity judgments \citep{kumar2022}. This growing divergence between task optimization and perceptual alignment motivates our investigation of whether alternative architectures and training paradigms, including Transformers trained with multimodal or self-supervised objectives, better preserve brain-like visual representations.

\subsection{CNN-Brain Correspondence via Spatial Neuroimaging}

A major advance in human neuroimaging came with \citet{kamitani2005}, who showed that multivoxel fMRI activity patterns in early visual cortex contain sufficient information to decode the orientation perceived or attended by a subject. This work demonstrated that distributed brain activity could be quantitatively decoded using multivoxel pattern analysis, laying the foundation for modern neural encoding and decoding research.

Deep neural networks later emerged as computational models of the visual hierarchy itself. \citet{yamins2014} showed that performance-optimized CNNs can predict neural responses across macaque visual cortex, with higher-level representations corresponding to responses in IT. \citet{cadieu2014} further demonstrated that high-performing CNNs can achieve representational performance comparable to primate IT cortex for core visual object recognition, while their high-level features can also predict individual IT neural responses. Together, these findings provided strong evidence that task-optimized CNNs capture important aspects of hierarchical visual representations in the primate ventral stream. However, both studies rely on invasive electrophysiology in
primates, recording a small number of IT/V4 neurons under static image
presentation; neither tests human EEG, nor any architecture beyond a
single CNN, leaving open whether the finding generalizes to non invasive
human recordings or to other model families. \citet{khaligh2014} extended this to human IT, via fMRI, and
macaque IT jointly, reporting that supervised CNNs match cortical
representations far better than unsupervised models. This comparison,
however, predates the vision transformer entirely: self-attention based
architectures, and the many alternative training objectives now
available for them, simply did not exist at the time, so the study could
not have asked whether the supervised training advantage it reports is a
property of CNNs specifically, or of visual hierarchies more generally.

Working with human fMRI directly, \citet{guclu2015}
showed that lower CNN layers correspond to early visual areas (V1--V3)
while deeper layers align with higher regions such as V4 and the lateral
occipital cortex. This mapping is built entirely on fMRI, using features
from a single pretrained AlexNet.
\citet{eickenberg2017} used voxel-wise encoding models with fMRI to show that representations from different CNN layers predict activity across multiple visual regions, with different layers exhibiting distinct correspondence across the visual hierarchy. However, their analysis relied on fMRI and a single CNN model, leaving open whether these layer–brain correspondences generalize across architectures and training objectives.

Building on earlier demonstrations of CNN–cortex correspondence \citep{yamins2014, khaligh2014}, \citet{schrimpf2018} proposed Brain-Score, a standardized benchmark that evaluates models using neural and behavioral predictivity. Their results showed that higher ImageNet accuracy does not necessarily imply greater brain-likeness. However, Brain-Score summarizes correspondence using the best-matching layer for each cortical region, rather than providing a layer-by-layer characterization of representational alignment throughout the network.

\subsection{CNN-Brain Correspondence via Temporal Neuroimaging}

\citet{cichy2016} combined fMRI and MEG through
representational similarity analysis and did resolve a temporal
dimension, showing a feedforward sweep from low level to category level
representations within roughly the first 200 ms. This is one of the few
studies to combined both modalities to obtain a spatio-temporal comparison, but it evaluates CNNs based architectures
exclusively. \citet{seeliger2018} extended earlier encoding studies by performing both encoding and decoding using MEG data. Using a pretrained VGG-S network, they demonstrated that CNN features accurately predict the spatiotemporal evolution of neural responses and that the learned encoding model can be inverted to decode the viewed object from MEG activity. However, the study was limited to a single CNN architecture (VGG-S). \citet{wen2018}
extended CNN to brain encoding to dynamic natural movie stimuli using
fMRI, validating hierarchical feature representations across visual
regions under naturalistic viewing, but movie fMRI still inherits fMRI's
coarse temporal resolution and again tests using single CNN architecture (Alexnet) only.

EEG captures the dynamics of visual processing at millisecond resolution, with neural representations evolving from relatively low-level visual features at earlier latencies toward increasingly abstract and categorical representations at later stages of processing \citep{grootswagers2019}. Despite this well-established
temporal structure, EEG-based encoding studies remain considerably fewer
than their fMRI counterparts, and most existing EEG model comparisons, as
reviewed above, evaluate only a single CNN architecture rather than a
matched set of diverse models. The THINGS-EEG2 dataset
\citep{gifford2022} provides a large-scale, high-SNR EEG resource
specifically designed for computational modeling of visual object
recognition, comprising responses from 10 participants across 16,540
training images and 200 test images. However, although the dataset has enabled large-scale
EEG encoding and decoding studies, it has not been used for a unified,
layer-wise comparison of a broad range of CNN and Transformer
architectures.

\subsection{Transformer-Brain Comparisons}

\citet{oota2022} directly compared CNNs, Vision Transformers, and multimodal models such as CLIP against human fMRI, making it the closest prior work to ours in evaluating both architecture families within a unified framework. However, the analysis remains limited to fMRI, whose coarse temporal resolution precludes tracking the millisecond by millisecond evolution of layer-wise representations or determining when CNNs and Transformers begin to diverge during visual processing.

Vision Transformers process images as sequences of patches using
multi-head self-attention, enabling each layer to model global spatial
relationships without the progressive spatial downsampling characteristic
of CNNs \citep{dosovitskiy2021}. \citet{conwell2022} performed the most
comprehensive controlled comparison of CNNs, Vision Transformers, and
other modern vision models to date, systematically isolating the effects
of architecture, training objective, and visual diet on brain predictivity.
Their results showed that, under matched training conditions, CNNs and
Vision Transformers achieve comparable alignment with the human ventral
visual system, while training objective and visual diet exert a greater
influence than architecture alone. The analysis relied exclusively
on fMRI and summarized each model using its best-predicting layer for
each cortical region, rather than characterizing how brain correspondence
evolves across all layers over time. Consequently, it remains unclear
whether CNNs and Vision Transformers diverge at specific stages of visual
processing or exhibit distinct layerwise temporal dynamics. This is the
question addressed in the present work.

\section{Methods}

\subsection{Dataset}

The study is conducted on the Alessandro T. Gifford dataset
\citep{gifford2022}, which contains EEG recordings from 10 participants
viewing natural object images. The experimental data is partitioned into
a training set comprising 16,540 unique images (presented with 4
repetitions each) and a test set comprising 200 images (presented with
80 repetitions each). While the EEG was originally recorded using a 63
channel system, the analysis focuses on vision related processing by
utilizing a subset of 17 posterior visual channels. The continuous neural
data are epoched from $-200$ to $800$ ms relative to stimulus onset and
downsampled to 100 time points, yielding a 10 ms temporal resolution.
Crucially, averaging the EEG responses over the high number of
repetitions, particularly the 80 repetitions in the test set, produces
neural responses with a high signal to noise ratio, making it an ideal
benchmark for evaluating models of human visual object recognition.

\subsection{Choice of Vision Models}

We test 8 models: 6 CNN based (AlexNet, VGG16, ResNet50, CORnet-S,
MoCo-v1, and CLIP-RN50) and 2 vision transformer based (DINOv2-ViT-B/14
and CLIP-ViT-B/32).

\subsubsection{Supervised CNNs} AlexNet \citep{krizhevsky2012}, VGG16
\citep{simonyan2015}, ResNet50 \citep{he2016}, and CORnet-S
\citep{kubilius2018}, all trained end to end on ImageNet classification.
Features were extracted at every named layer from the first convolutional
output to the final classification layer.

\subsubsection{Self-supervised CNN} MoCo-v1 \citep{he2020}, a ResNet-50
backbone trained with a contrastive instance discrimination objective and
no class labels, matching the supervised CNNs in architecture but not
objective. Layers are extracted identically to the supervised CNNs.

\subsubsection{Vision Transformers} DINOv2-ViT-B/14 \citep{oquab2023} is
trained via self-distillation between a student and a slowly updated
teacher network, with no labels and no language supervision.
CLIP-ViT-B/32 \citep{radford2021} is trained to align images with natural
language captions. Features for both are extracted at the output of each
of the 12 transformer blocks.

\subsubsection{Architecture and Objective Controlled Comparisons} CLIP-RN50
shares CLIP-ViT-B/32's contrastive language-image training objective but
uses a ResNet-50 backbone instead of a transformer. Together, MoCo-v1
(architecture fixed, objective changed) and CLIP-RN50 (objective fixed,
architecture changed) let us ask whether the deep layer divergence
reported in Section~\ref{results} tracks architecture, training
objective, or both. All eight models are used as frozen feature
extractors; no fine tuning was performed.

\subsection{EEG Encoding Model}

Brain responses are predicted through a fixed pipeline: features are
reduced via Kernel PCA, mapped to EEG via OLS regression, and evaluated
by correlating predictions against held-out test EEG. This pipeline is
applied both per layer and on all layers concatenated, described in
items 3 and 4 below.

\subsubsection{Kernel PCA Dimensionality Reduction} This step follows the
pipeline $\mathbf{X}_{\text{raw}} \to \text{standardize} \to
\text{KernelPCA} \to \mathbf{X}$. Raw feature dimensionality
ranges from a few thousand (single CNN layers) to hundreds of thousands
(all layer ViT concatenation). Features are first standardized to zero
mean and unit variance, then reduced to $P{=}3{,}000$ components via
polynomial kernel PCA:

\begin{equation}
    K(\mathbf{x}, \mathbf{y}) = (\gamma\,\mathbf{x}^{\!\top}\mathbf{y} + 1)^{4},
    \quad \gamma = \tfrac{1}{D}
\end{equation}

where $D$ is the standardized input dimensionality and $\gamma{=}1/D$,
coefficient 1, and degree 4 match the standard KernelPCA defaults. The
kernel matrix is fitted and eigendecomposed once on the full training
features; test features are projected out of sample using the fitted
eigenvectors and eigenvalues, without ever recomputing the decomposition
on test data. This nonlinear projection captures feature interactions up
to degree 4 while keeping the downstream regression tractable.

\begin{figure}[pos=htbp]
\centering
\includegraphics[width=0.5\linewidth]
    {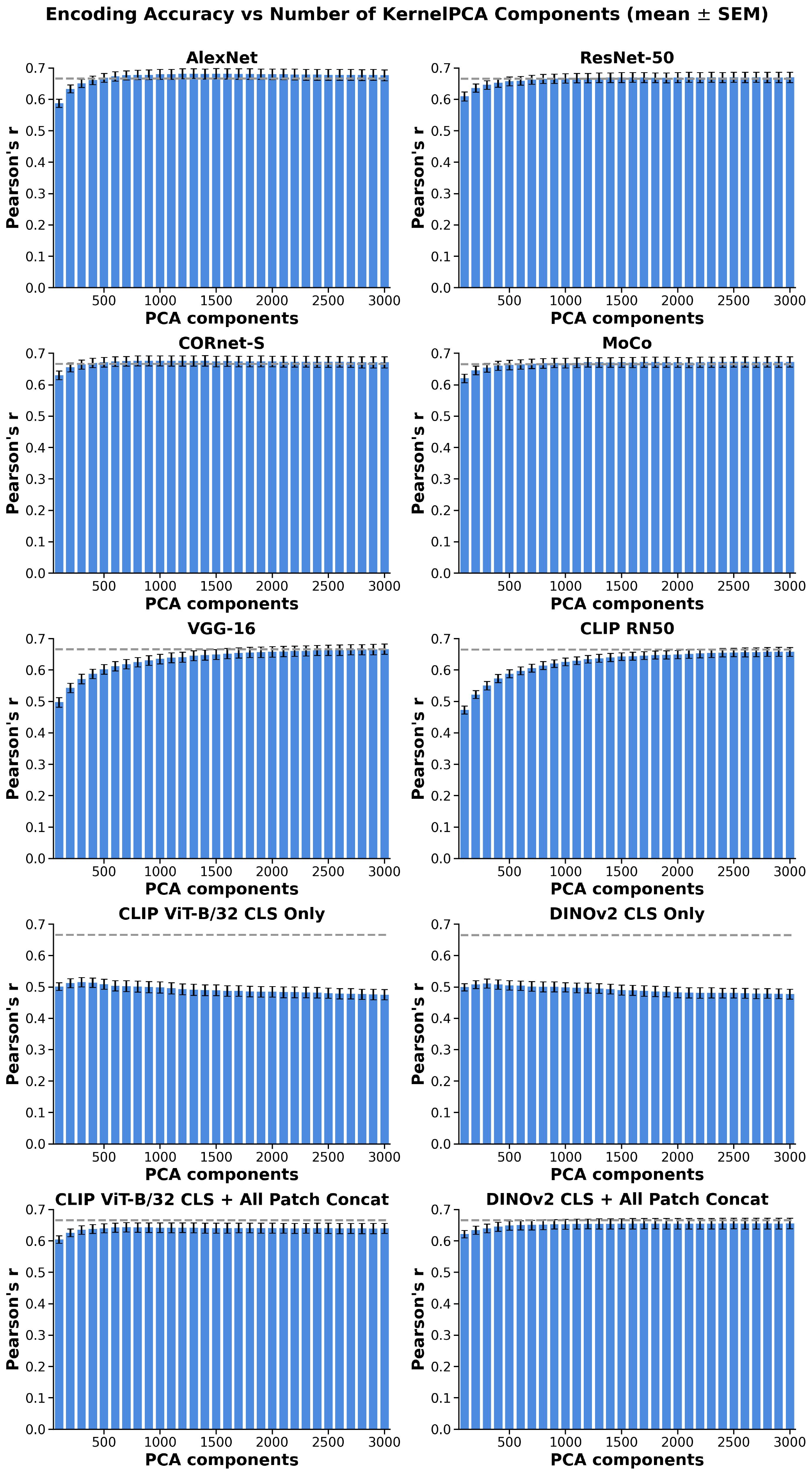}
\caption{Whole model EEG encoding accuracy (Pearson $r$, mean $\pm$ SEM
over 10 subjects) versus number of retained KernelPCA components, swept
up to 3,000, for each of the eight base models. Dashed line: noise
ceiling lower bound.}
\label{fig:pca_sweep}
\end{figure}

\textbf{Choice of component count.} The value 3,000 was not chosen
arbitrarily. Fig.~\ref{fig:pca_sweep} sweeps the number of retained
KernelPCA components from a 100 to 3,000, for each of the
eight base models, using the same full time window peak metric used
throughout the paper. Six of the eight models plateau well before 2,000
components, with negligible change (within $\pm0.003$ in Pearson $r$)
from 2,000 to 3,000. VGG-16 and CLIP-RN50 continue rising slightly
beyond 2,000, but the gain from 2,000 to 3,000 components is small
(under 0.01 in $r$ for both), indicating that 3,000 components is
sufficient to capture the vast majority of the achievable encoding
accuracy across all eight models.

\subsubsection{Ordinary Least-Squares Encoding} This step follows the
pipeline $\mathbf{X}, \mathbf{Y}_\text{train} \to \text{OLS} \to
\mathbf{W}$. The EEG training target for
each image is the response averaged across all available training
repetitions. Encoding weights $\mathbf{W}$ are estimated via ordinary
least squares on the training set:

\begin{equation}
    \mathbf{W} =
    (\mathbf{X}_\text{train}^{\!\top}\mathbf{X}_\text{train})^{-1}
    \mathbf{X}_\text{train}^{\!\top}\mathbf{Y}_\text{train}
\end{equation}

where $\mathbf{X}_\text{train} \in \mathbb{R}^{N_\text{train} \times P}$ is
the projected training feature matrix and $\mathbf{Y}_\text{train}$ is the
EEG response matrix. Weights are estimated separately for each subject.

Here, $N_\text{train}$ is the number of training images, $P$ is the
number of retained KernelPCA components, $C$ is the
number of EEG channels, and $T$ is the number of timepoints. The EEG response tensor is flattened along the channel and
time axes before regression, so that $\mathbf{Y}_\text{train} \in
\mathbb{R}^{N_\text{train} \times (C \times T)}$.

\subsubsection{Layer-wise Encoding} Each layer follows the pipeline
$\mathbf{X}_l \to \text{KernelPCA}(\mathbf{X}_l) \to \text{OLS} \to
\hat{\mathbf{Y}}_l \to r_l$, for all $l \in \{1, \dots, L\}$. In this mode, each layer
$l \in \{1, \dots, L\}$ is analyzed independently: its raw feature matrix
$\mathbf{X}_l \in \mathbb{R}^{N \times D_l}$, where $N$ is the number of
images and $D_l$ is the raw feature dimensionality of layer $l$, is passed
through its own KernelPCA reduction to
$\mathbf{X}_l' \in \mathbb{R}^{N \times P}$, and a separate OLS encoding
model $\mathbf{W}_l$ is fit and evaluated for that layer alone. This is
what produces the per-layer encoding scores reported throughout
Section~\ref{results}.

\subsubsection{All Layer Concatenated Encoding} This mode follows the
pipeline $\mathbf{X}_{\text{concat}} = \mathbf{X}_1 \,\|\, \mathbf{X}_2
\,\|\, \cdots \,\|\, \mathbf{X}_L \to \text{KernelPCA} \to \text{OLS}
\to \hat{\mathbf{Y}} \to r$. In this mode, every layer's
raw features are concatenated along the feature axis before reduction:

\begin{equation}
\mathbf{X}_{\text{concat}} = \big[\, \mathbf{X}_1 \,\|\, \mathbf{X}_2 \,\|\,
\cdots \,\|\, \mathbf{X}_L \,\big] \in \mathbb{R}^{N \times \sum_{l=1}^{L} D_l}
\end{equation}

where $L$ is the total number of layers in the model and $\|$ denotes
concatenation along the feature dimension. A single KernelPCA reduction is
then fitted on $\mathbf{X}_{\text{concat}}$ to obtain
$\mathbf{X} \in \mathbb{R}^{N \times P}$, and a single OLS model
$\mathbf{W}$ is fit on this combined representation, producing predicted
responses over $\mathbb{R}^{N \times (C \times T)}$. This
is the whole model score used for the overall model ranking
(Table~\ref{tab:ranking}).

\subsection{Evaluation and Statistical Procedure}

\subsubsection{Test set Prediction} Once $\mathbf{W}$ is estimated, the
model's predicted EEG response on the test set,
$\hat{\mathbf{Y}} \in \mathbb{R}^{N_\text{test} \times C \times T}$, is
computed once from the projected test features and held fixed for all
subsequent evaluation steps.

\subsubsection{Split half Correlation} Brain model alignment is measured as
the Pearson correlation between the fixed predicted response and actual
EEG on the test set, computed independently for each channel $c$ and
timepoint $t$. Rather than compare the prediction against a single fixed
average of the test repetitions ($R_\text{test}$), we follow the split
half procedure of \citet{gifford2022}: for each of
$i = 1, \dots, M$ random splits, the $R_\text{test}$ repetitions are
divided into two halves and one half is averaged into $y^{(1),i}_{c,t}$.
The correlation for that split is

\begin{equation}
r^{(i)}_{c,t} =
\frac{\sum_{n=1}^{N_\text{test}} (\hat{y}_{n,c,t} - \bar{\hat{y}}_{c,t})
(y^{(1),i}_{n,c,t} - \bar{y}^{(1),i}_{c,t})}
{\sqrt{\sum_{n=1}^{N_\text{test}} (\hat{y}_{n,c,t} - \bar{\hat{y}}_{c,t})^2}\
\sqrt{\sum_{n=1}^{N_\text{test}} (y^{(1),i}_{n,c,t} - \bar{y}^{(1),i}_{c,t})^2}}
\end{equation}

where $N_\text{test}$ is the number of test images, $\hat{y}_{n,c,t}$
is the model's predicted response, and the sums and means are taken over
test images $n$. The reported correlation averages over all $M{=}100$
splits,

\begin{equation}
r_{c,t} = \frac{1}{M}\sum_{i=1}^{M} r^{(i)}_{c,t},
\end{equation}

which is less sensitive to which particular trials happen to fall into a
given half than a single split would be.

\subsubsection{Temporal and Layer-wise Peak Encoding} This channel wise
correlation is then averaged over all $C$ channels at each of the $T$
time points spanning the full $-200$ to $800$ ms time window. For each
layer and subject, we take the peak value of this channel averaged
correlation over the entire time window. We do not restrict this peak
search to a sub window: the full time window is searched so that no
processing stage is favored by the choice of time range.

\subsubsection{Noise Ceiling} Since trial to trial EEG variability caps how
high any model's correlation can go, we bound it using the same splits:
the \textit{lower bound} is
$\mathrm{Corr}(y^{(1),i}_{c,t}, y^{(2),i}_{c,t})$ between the two
independent halves, and the \textit{upper bound} is
$\mathrm{Corr}(y^{(1),i}_{c,t}, \bar{y}_{c,t})$ between one half and the
full $R_\text{test}$ repetition average, both averaged over the
$M$ iterations.

\subsubsection{Subject level Aggregation} The per layer, per subject peak
value obtained above is then averaged across all  subjects to obtain a
single encoding score for each layer or model.

\FloatBarrier
\section{Experimental Evaluation and Discussion}
\label{results}

We evaluate brain correspondence across vision models along four
complementary axes. We begin by examining how encoding accuracy evolves
across the depth of each network, layer by layer. We then investigate
how different strategies for aggregating transformer tokens (CLS,
patch, and their combinations) influence encoding performance. Next,
we compare models at an aggregate level correspondence with brain activity.
Finally, we present a topography study of encoding accuracy across EEG
channels to identify where these effects are strongest.

\subsection{Layer-wise Encoding Across CNN Models}

We first examine the correspondence between CNN representations and EEG
responses as a function of network depth, evaluating each layer
independently rather than aggregating layers into broader stages. For
each layer, the corresponding EEG response is predicted using the
encoding procedure described in Section III, and the Pearson
correlation between predicted and measured responses is evaluated
across the visual EEG channels and the complete $-200$ to $800$ ms
time window.

\begin{figure}[pos=htbp]
\centering
\includegraphics[width=1\linewidth]
    {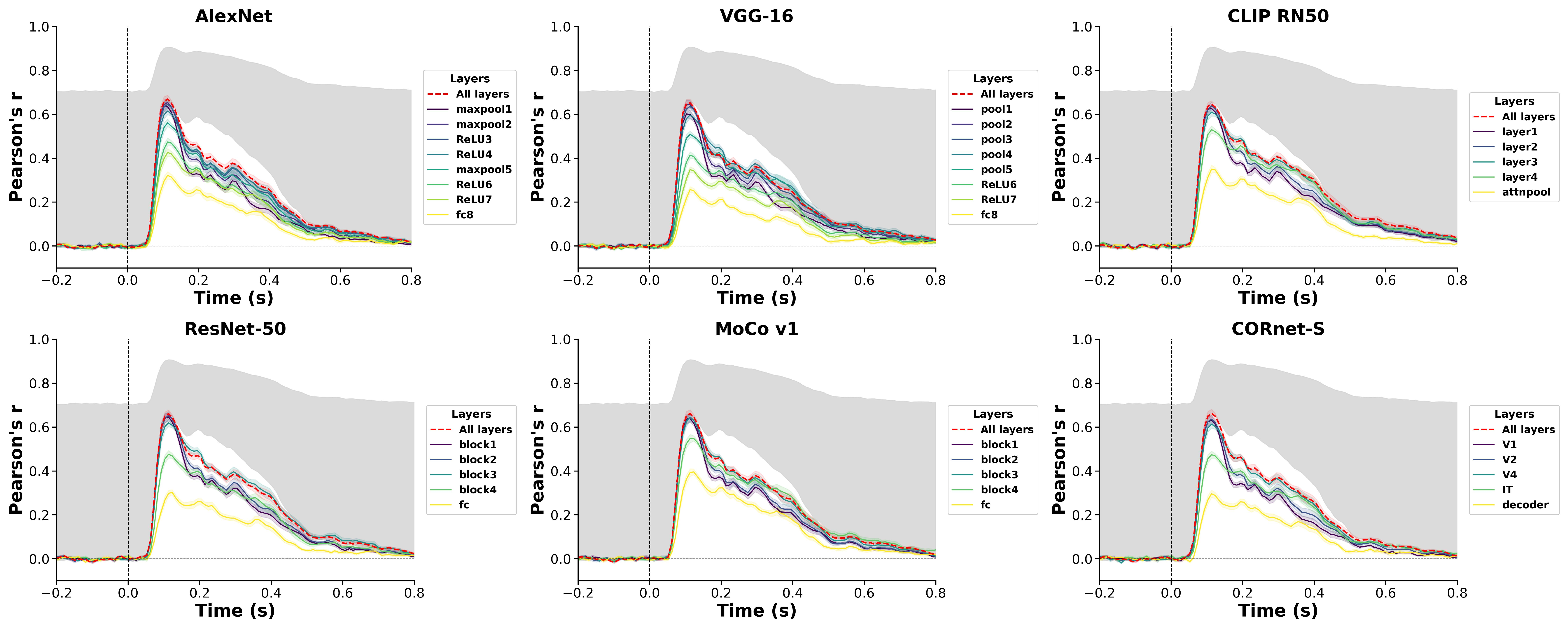}
\caption{Layer-wise temporal encoding curves for the CNN based models
(AlexNet, VGG-16, CLIP-RN50, Resnet 50, MoCo-v1, and CORnet-S), each panel showing
Pearson $r$ across the full $-200$ to $800$ ms epoch for every layer of
the respective model.}
\label{fig:panel_other_cnns}
\end{figure}

Fig.~\ref{fig:panel_other_cnns} shows the layer-wise temporal encoding
profiles for the CNN based models. Across the different CNN
architectures, the earliest layers generally produce strong
correspondence with the measured EEG responses: peak layer-wise
accuracy is consistently reached within 60--150 ms window,
with AlexNet's \texttt{maxpool2}, ResNet-50's \texttt{block2}, and
MoCo-v1's \texttt{block2} reaching the highest values ($r=0.654$,
$0.653$, and $0.646$ respectively), while CORnet-S's
\texttt{V2} marks the lower end of this range at $r=0.636$
(Table~\ref{tab:layer_peak}). For example, early layers such as
AlexNet's \texttt{maxpool1} and \texttt{maxpool2}, VGG-16's
\texttt{pool1} and \texttt{pool2}, Resnet 50 \texttt{block 1}
CLIP-RN50's \texttt{layer1}, MoCo-v1's first residual block, and the
\texttt{V1} and \texttt{V2} layers of CORnet-S exhibit strong encoding
responses. Their temporal profiles show a pronounced increase in
encoding accuracy shortly after stimulus onset, with the strongest
responses occurring during the early portion of the visual response.

\begin{table}[htbp]
\caption{Each model's peak layer-wise encoding accuracy (Pearson $r$),
achieved within the 60--150 ms post-stimulus window.}
\label{tab:layer_peak}
\begin{tabular*}{\tblwidth}{@{}LLL@{}}
\toprule
\textbf{Model} & \textbf{Peak Layer} & \textbf{Peak $r$} \\
\midrule
AlexNet                            & maxpool2 & 0.654 \\
VGG-16                             & pool2    & 0.646 \\
ResNet-50                          & block2   & 0.653 \\
MoCo-v1                            & block2   & 0.646 \\
CORnet-S                           & V2       & 0.636 \\
CLIP-RN50                          & layer2   & 0.638 \\
CLIP-ViT-B/32 (CLS only)           & block\_3 & 0.448 \\
CLIP-ViT-B/32 (Mean patch)         & block\_4 & 0.459 \\
CLIP-ViT-B/32 (CLS+Mean patch)     & block\_4 & 0.473 \\
CLIP-ViT-B/32 (CLS+All patch)      & block\_2 & 0.629 \\
DINOv2-ViT-B/14 (CLS only)         & block\_3 & 0.448 \\
DINOv2-ViT-B/14 (Mean patch)       & block\_2 & 0.467 \\
DINOv2-ViT-B/14 (CLS+Mean patch)   & block\_3 & 0.475 \\
DINOv2-ViT-B/14 (CLS+All patch)    & block\_2 & 0.660 \\
\bottomrule
\end{tabular*}
\end{table}

This early correspondence is consistent with the hierarchical
organization of the human ventral visual stream, in which early visual
regions such as V1 and V2 are associated with the representation of
relatively low level visual properties such as edges, contours,
textures, and basic shape information. The strong encoding observed in
the early and intermediate CNN layers therefore indicates that, at these
stages, the CNN representations follow the expected progression of
visual processing observed in the brain. In particular, the
correspondence between early CNN representations and the early EEG
response is consistent with the rapid feedforward processing of low
level visual information during the initial stages of visual
perception.

\begin{figure}[pos=htbp]
\centering
\includegraphics[width=1\linewidth]
    {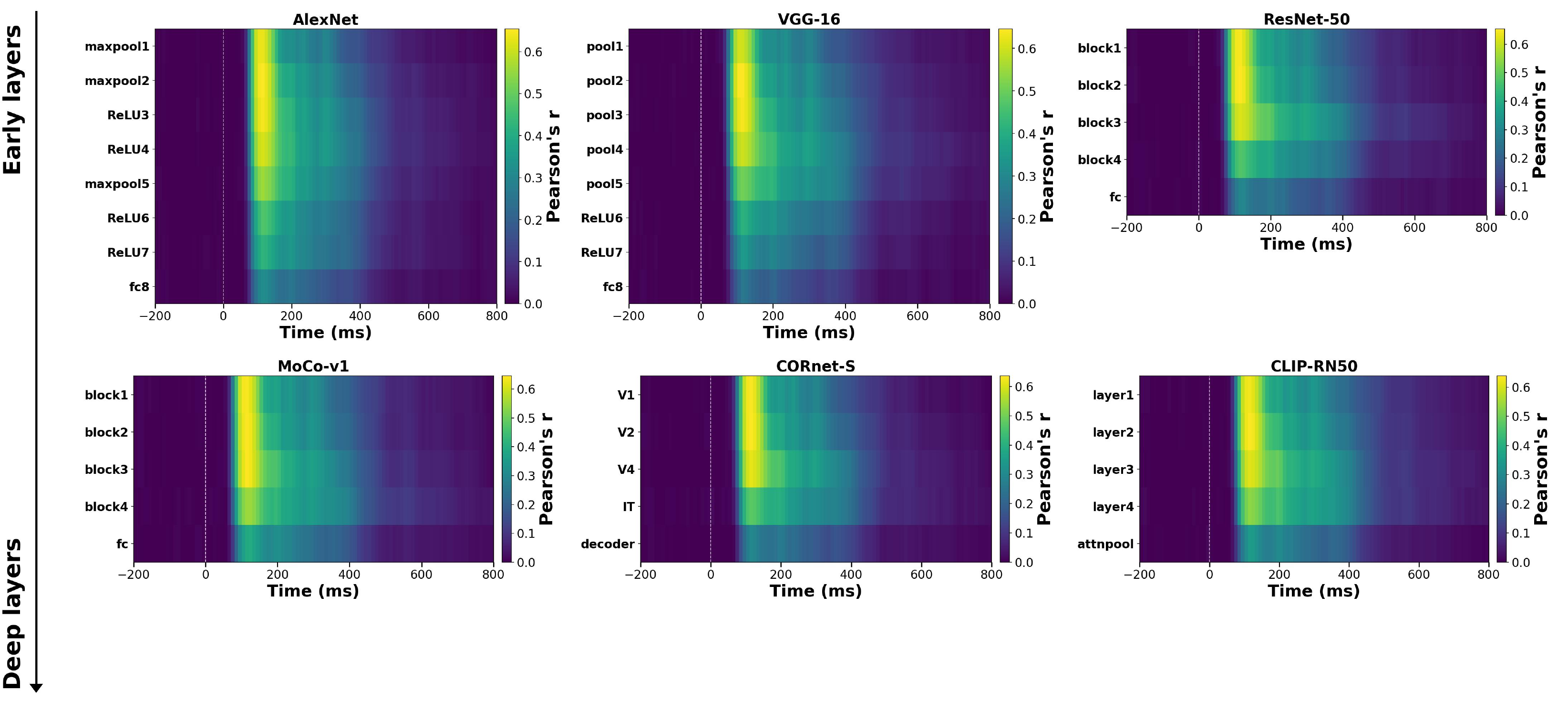}
\caption{Layer $\times$ time EEG encoding heatmaps for the CNN based models
(AlexNet, VGG-16, ResNet-50, MoCo-v1, CORnet-S, and CLIP-RN50), showing
Pearson $r$ across the full $-200$ to $800$ ms epoch (x-axis) and network
depth (y-axis, early to deep layers).}
\label{fig:panel_cnn_heatmaps}
\end{figure}

As the representations progress toward deeper layers, the encoding
profiles generally become weaker. This reduction is particularly evident
in the final layers of the CNNs, including AlexNet's \texttt{fc8},
VGG-16's \texttt{fc8}, CLIP-RN50's attention pooling representation,
MoCo-v1's final fully connected representation, and CORnet-S's
\texttt{IT} layer. This pattern is visible directly in the layer $\times$ time heatmaps of Fig.~\ref{fig:panel_cnn_heatmaps}: the brightest, strongest correlation values are concentrated in the top rows (earliest layers) shortly after stimulus onset, while the bottom rows (deepest layers) remain visibly darker throughout the time window, indicating consistently weaker correspondence with the EEG response. These deeper representations show weaker correspondence with the EEG response than the earlier layers, particularly after the initial visual response. Although the CNNs
provide strong encoding at shallow and intermediate depths that is
consistent with the early stages of the visual hierarchy, this
correspondence with later layers does not remain equally strong as processing progresses
toward the higher level and semantic representations.

The same pattern is observed across the different CNN
training settings. In particular, the self-supervised MoCo-v1
representation follows a layer-wise trajectory broadly similar to the
supervised CNNs, while CLIP-RN50 also exhibits a reduction in encoding
toward its later representation. CORnet-S provides an additional
reference because its layers are explicitly labeled according to stages
of the visual hierarchy; nevertheless, its final \texttt{IT}
representation also exhibits weaker EEG encoding than its earlier
representations. This is notable because the \texttt{IT} label of
CORnet-S was explicitly designed to correspond to a high level visual
processing stage, yet its final representation does not maintain the
strong EEG correspondence observed in the earlier CNN layers.

\FloatBarrier

\FloatBarrier

\subsection{Block-wise Encoding of Vision Transformers}

We next examine the corresponding layer-wise encoding profiles of
CLIP-ViT-B/32 and DINOv2-ViT-B/14. Unlike the CNNs, the transformer
models do not show a simple reduction in encoding accuracy with
increasing depth, using the default CLS token representation, both
reach their peak correlation at \texttt{block\_3} ($r=0.448$ for each),
well below the $r=0.636$--$0.654$ achieved by the strongest CNN layers
(Table~\ref{tab:layer_peak}). Instead, their block-wise responses
exhibit a progressive change in temporal encoding across transformer
depth. The later blocks retain relatively strong correspondence with
the EEG response, particularly after 200ms as shown in
Figs.~\ref{fig:panel_dinov2_variants} and \ref{fig:panel_clip_vit_variants}. In this respect, the transformer representations exhibit a
depth dependent progression that differs from the reduction observed in
the deepest CNN representations. This pattern is visible directly in the
block $\times$ time heatmaps of Fig.~\ref{fig:panel_dinov2_heatmaps} and
Fig.~\ref{fig:panel_clip_vit_heatmaps}: unlike the CNN heatmaps, the
bottom rows (deepest blocks) remain bright well into the later portion of
the time window rather than fading, showing that both DINOv2-ViT-B/14 and
CLIP-ViT-B/32 sustain strong correspondence with the EEG response at
depth.

This depth dependent difference raises a question about the
representation used for the transformer models. CNN layers retain a
spatial feature map, whereas vision transformers represent an image as a
sequence of a global CLS token and spatial patch tokens, so the
representation supplied to the encoding model depends on how information
across these tokens is aggregated. We therefore examine whether the
observed transformer encoding pattern depends on the token
representation used.

\subsubsection{Transformer Token Representation Analysis}

Vision transformers represent an image as a sequence of tokens,
consisting of a global CLS token and spatial patch tokens. Unlike
convolutional feature maps, this representation allows the spatial patch
information to be either retained or compressed before being used for
EEG encoding. To determine how this representational choice affects
brain encoding, we evaluate four token aggregation strategies for both
DINOv2-ViT-B/14 and CLIP-ViT-B/32.

\begin{figure}[pos=htbp]
\centering
\includegraphics[width=0.68\linewidth]
    {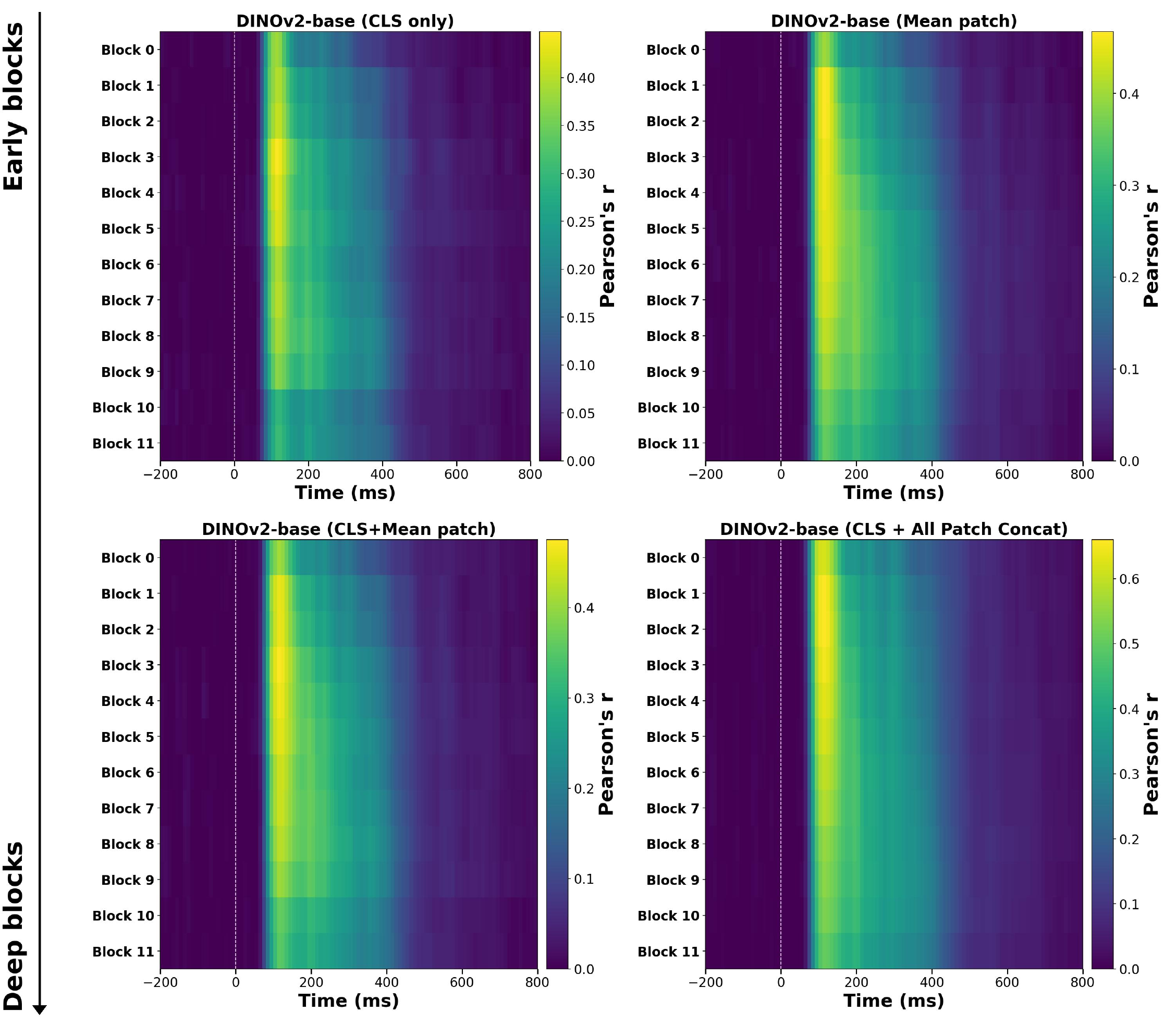}
\caption{Block $\times$ time EEG encoding heatmaps for DINOv2-ViT-B/14's
four token aggregation variants (CLS token only, mean of patch tokens,
CLS + mean of patch tokens, CLS + all patch tokens concatenated).}
\label{fig:panel_dinov2_heatmaps}
\end{figure}

\begin{figure}[pos=htbp]
\centering
\includegraphics[width=0.68\linewidth]
    {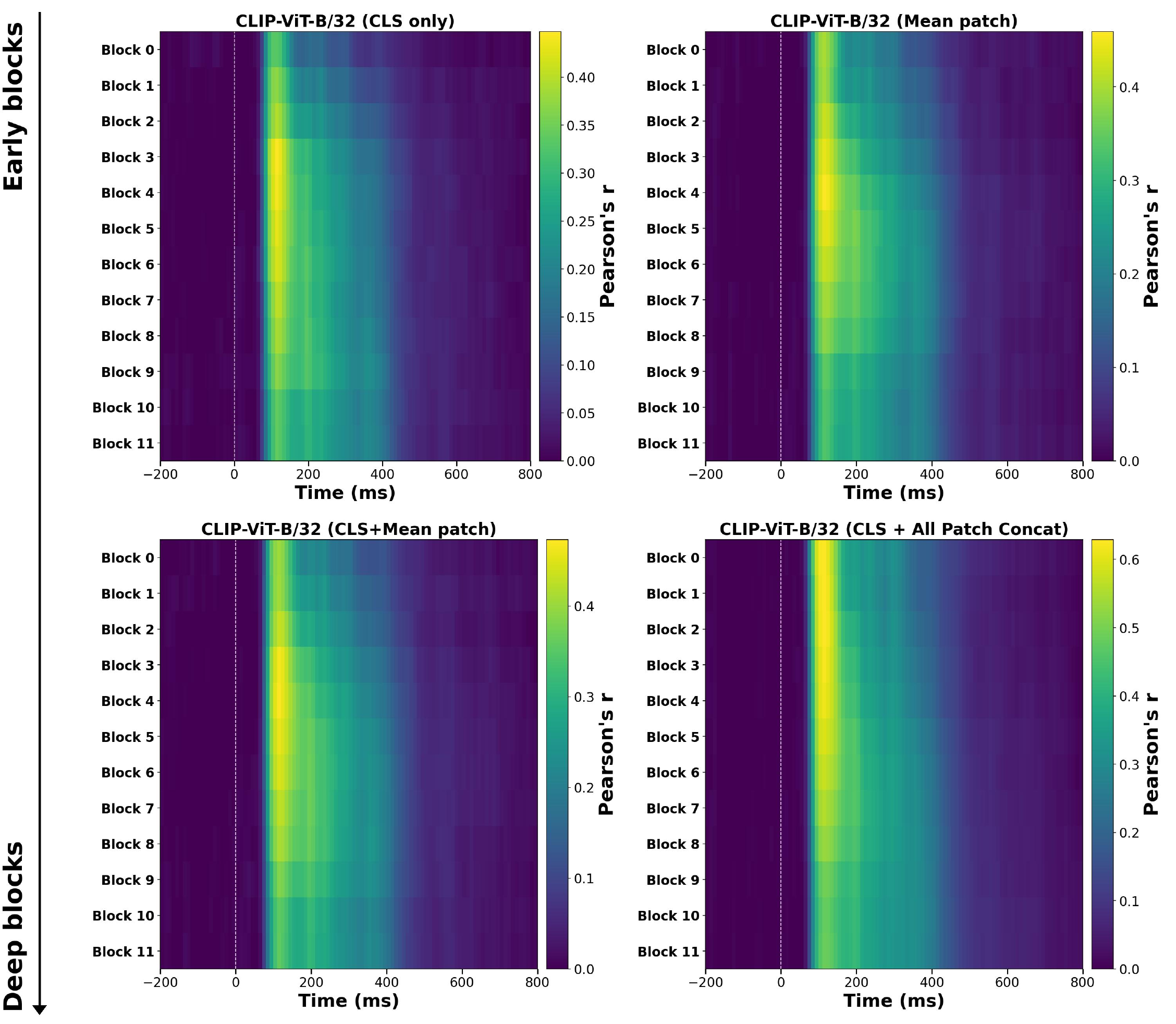}
\caption{Block $\times$ time EEG encoding heatmaps for CLIP-ViT-B/32's
four token aggregation variants (CLS token only, mean of patch tokens,
CLS + mean of patch tokens, CLS + all patch tokens concatenated).}
\label{fig:panel_clip_vit_heatmaps}
\end{figure}

The four representations are: (1) the CLS token alone, (2) the mean of
all patch tokens, (3) the CLS token concatenated with the mean patch
representation, and (4) the CLS token concatenated with all individual
patch tokens. The first three strategies compress the spatial patch
sequence into a global or pooled representation, whereas the fourth
retains the individual spatial patch representations without pooling.
This same advantage of the CLS + all patches representation is visible in
the block $\times$ time heatmaps of Fig.~\ref{fig:panel_dinov2_heatmaps}
and Fig.~\ref{fig:panel_clip_vit_heatmaps}, which show consistently
brighter, stronger correlation across blocks and time compared with the
pooled token variants.

\subsubsection{Temporal Encoding Across Token Representations}

\begin{figure}[pos=htbp]
\centering
\includegraphics[width=0.83\linewidth]
    {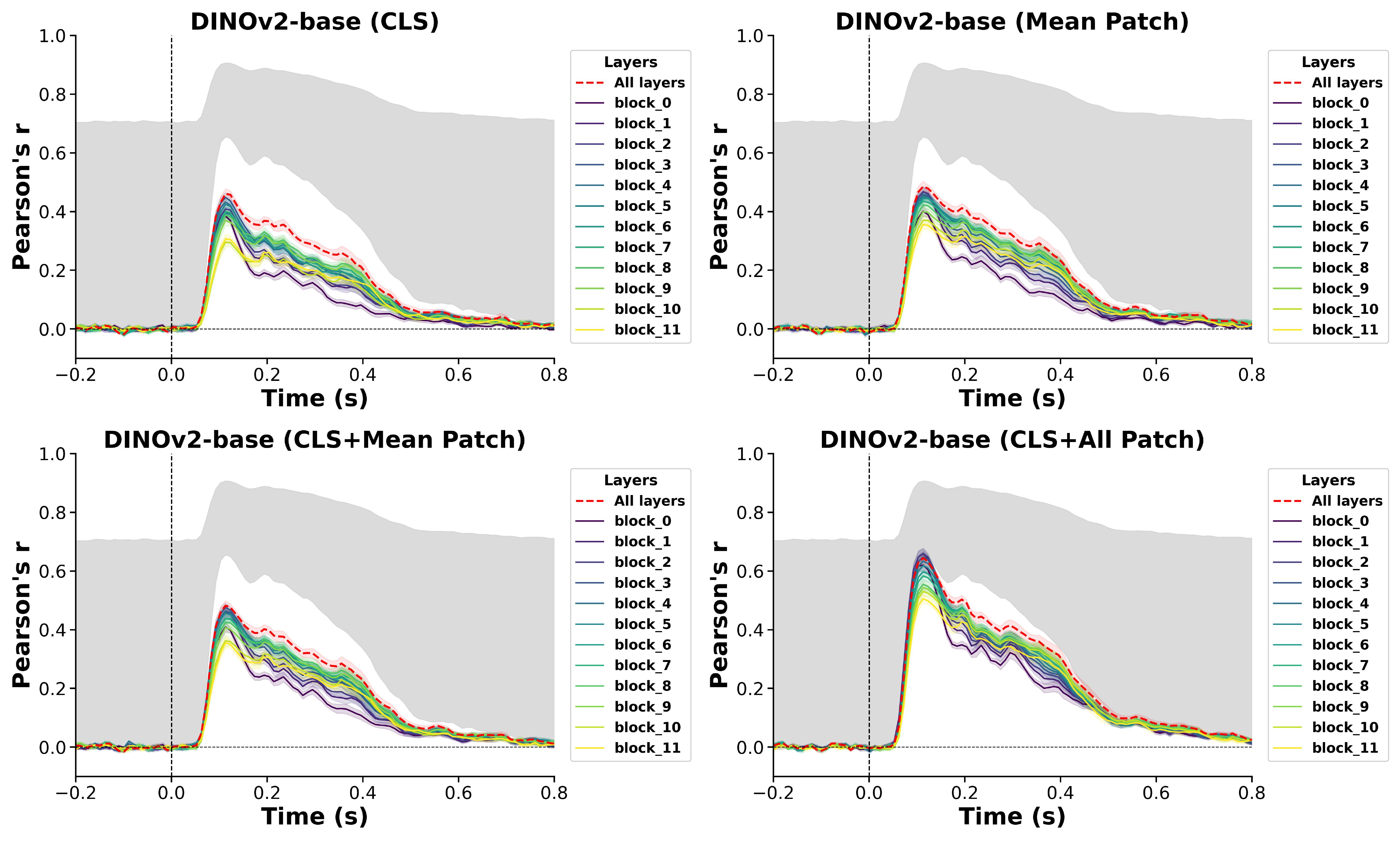}
\caption{Layer-wise temporal encoding curves for DINOv2-ViT-B/14's four
token aggregation variants (CLS token only, mean of patch tokens, CLS +
mean of patch tokens, CLS + all patch tokens concatenated).}
\label{fig:panel_dinov2_variants}
\end{figure}

\begin{figure}[pos=htbp]
\centering
\includegraphics[width=0.83\linewidth]
    {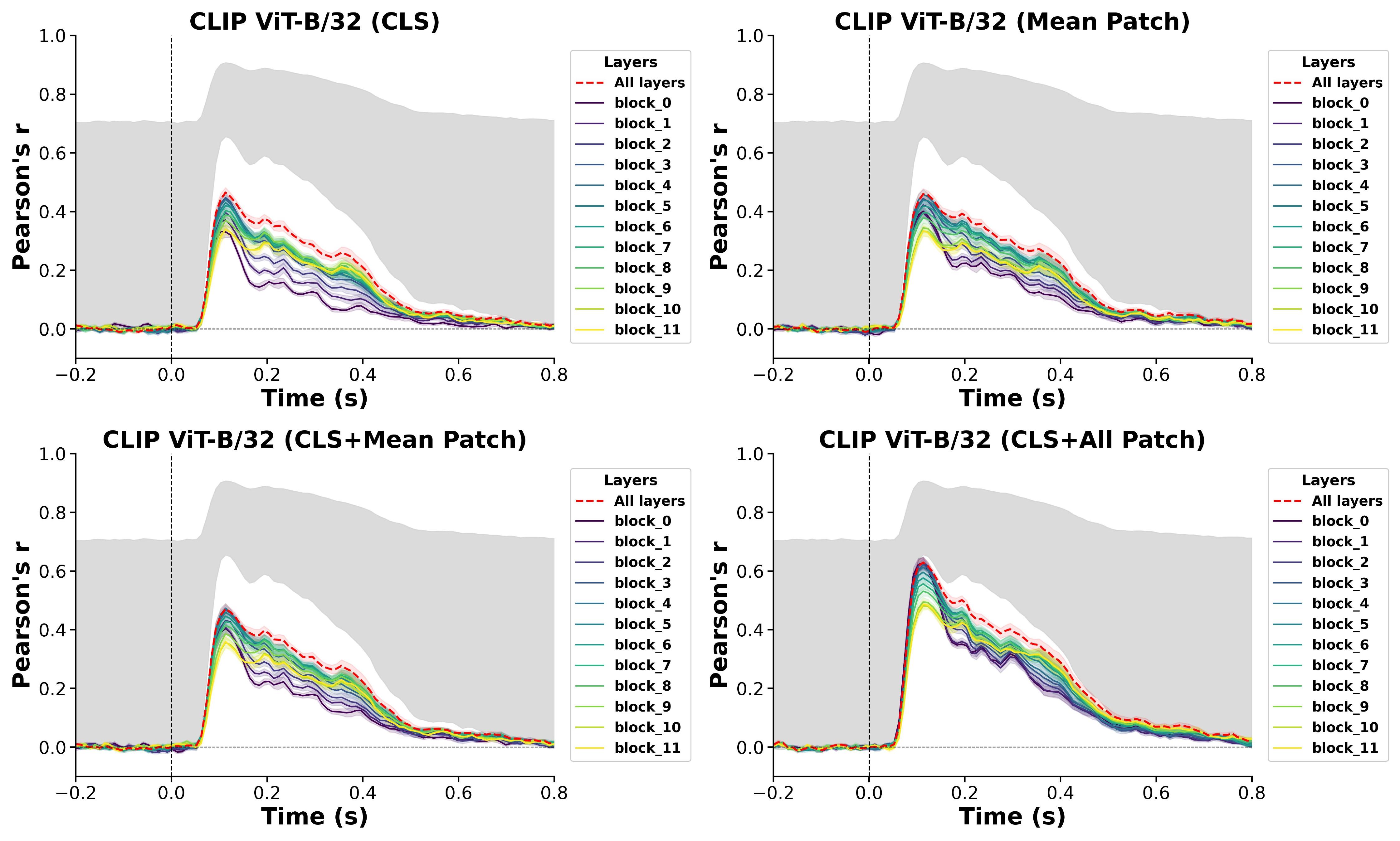}
\caption{Layer-wise temporal encoding curves for CLIP-ViT-B/32's four
token aggregation variants (CLS token only, mean of patch tokens, CLS +
mean of patch tokens, CLS + all patch tokens concatenated).}
\label{fig:panel_clip_vit_variants}
\end{figure}

Figs.~\ref{fig:panel_dinov2_variants} and \ref{fig:panel_clip_vit_variants}
show the layer-wise temporal encoding profiles for the four token
representations of DINOv2-ViT-B/14 and CLIP-ViT-B/32, respectively.
Across both transformer models, the choice of token representation
produces a substantial difference in encoding accuracy. The CLS only,
mean patch, and CLS+ mean representations show broadly similar
temporal profiles, whereas the CLS+all patches concatenated representation
consistently produces stronger encoding responses.

\FloatBarrier

\FloatBarrier

The CLS+all patches concatenated representation maintains stronger EEG encoding during the later post-stimulus response than the pooled representations, indicating a sustained advantage across the measured time window.

Importantly, retaining the individual patch tokens strengthens the
correspondence at the earlier stages of processing. This suggests that
the weaker early encoding observed with the native or pooled transformer
representations is not necessarily a limitation of the transformer
itself, but is strongly influenced by which components of the
internal token representation are provided to the encoding model.

\subsubsection{Effect of Token Aggregation on Encoding Accuracy}

\begin{figure}[pos=htbp]
\centering
\includegraphics[width=0.7\linewidth]
    {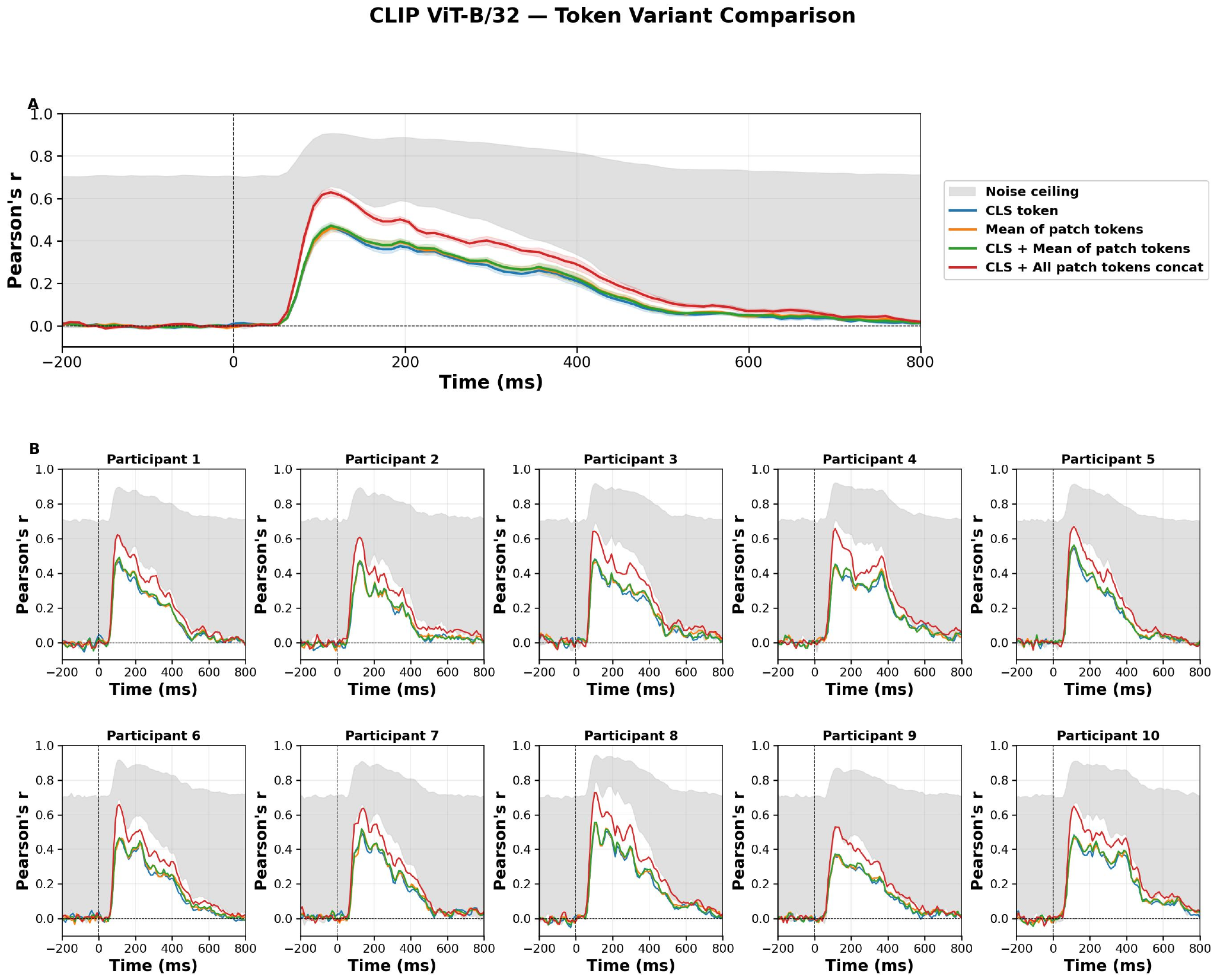}
\caption{Temporal EEG encoding profiles for CLIP-ViT-B/32 across four token representations. Panel A: group average across 10 participants; Panel B: individual participant profiles.}
\label{fig:clip_variants}
\end{figure}

\begin{figure}[pos=htbp]
\centering
\includegraphics[width=\linewidth]
    {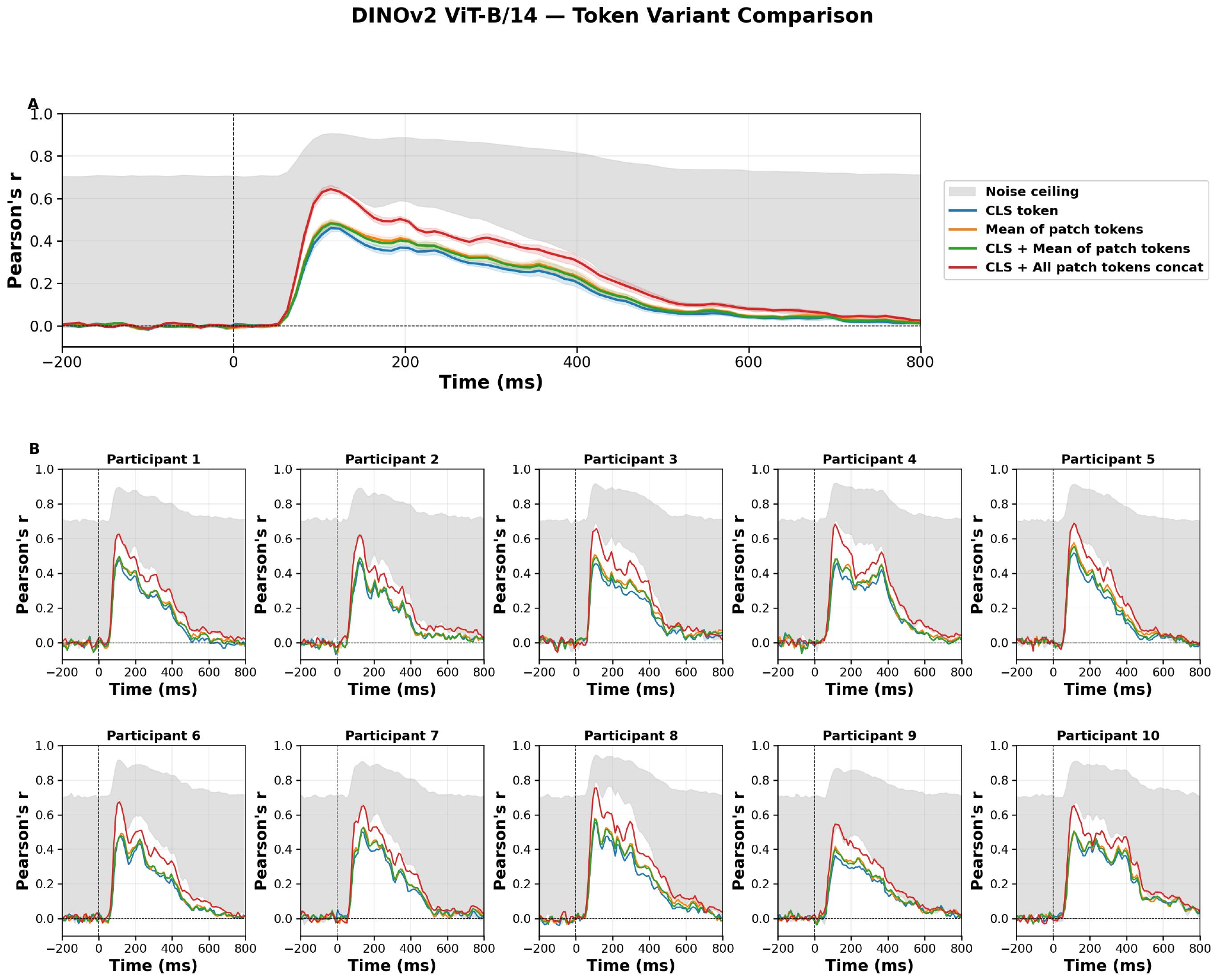}
\caption{Temporal EEG encoding profiles for DINOv2-ViT-B/14 across four token representations. Panel A: group average across 10 participants; Panel B: individual participant profiles.}
\label{fig:dinov2_variants}
\end{figure}

\begin{figure}[pos=htbp]
\centering
\includegraphics[width=0.7\linewidth]
    {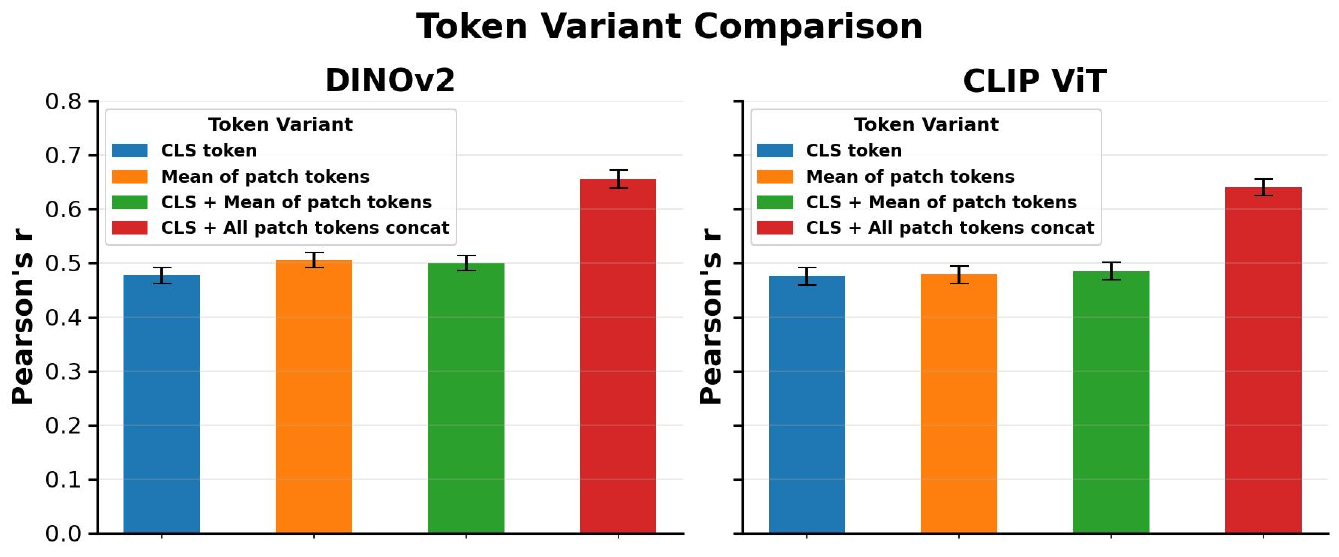}
\caption{Peak EEG encoding accuracy across transformer token representations.}
\label{fig:token_variant_comparison}
\end{figure}

To summarize the differences observed in the temporal encoding profiles,
Figs.~\ref{fig:clip_variants} and \ref{fig:dinov2_variants} compare the
four token representations for CLIP-ViT-B/32 and DINOv2-ViT-B/14,
respectively. In both figures, Panel A shows the group averaged temporal
encoding curves across all 10 participants, while Panel B shows the
corresponding temporal profiles for each participant individually
(Participants 1--10). Across both transformer models, the three
representations that compress the spatial patch sequence, CLS only,
mean of patch tokens, and CLS + mean of patch tokens, show substantially
lower encoding accuracy than the representation that preserves the
individual patch tokens. In the group averaged profiles shown in Panel A,
the CLS + all patches concatenated representation consistently produces the strongest
encoding response and maintains a clear advantage over the pooled
representations throughout the post-stimulus period. The same pattern is
also visible across the individual participant profiles in Panel B,
although the magnitude and temporal shape of the response vary across
participants. For DINOv2, the three pooled representations achieve peak
correlations of approximately $r{=}0.48$--$0.51$, whereas the
CLS + all patches representation reaches approximately $r{=}0.66$. A
similar pattern is observed for CLIP-ViT, where the three pooled
representations remain near $r{=}0.48$, while the CLS + all patches
representation reaches $r{=}0.64$. The difference between the
CLS + all patches representation and the pooled representations is
substantially larger than the differences among the three pooled
representations themselves. These results indicate that the dominant
factor is whether the individual spatial patch tokens are retained,
rather than which particular pooled representation is used. Preserving
the complete patch-level representation therefore provides substantially
stronger EEG encoding for both transformer models. Fig.~\ref{fig:token_variant_comparison}
summarizes this peak encoding accuracy across all four token
representations for both transformer models.

\FloatBarrier

\subsection{Model-Level Performance and Architecture/Training Objective Controls}

The layer-wise analysis revealed a qualitative difference between the
two architecture families: CNNs show their strongest correspondence at
early layers, with encoding accuracy declining toward deeper layers,
whereas transformers maintain comparatively stronger correspondence into
deeper blocks, particularly during the later post-stimulus period
(beyond approximately 200 ms) associated with higher-level visual
regions such as V4 and IT. The token analysis further showed that this
transformer advantage depends on preserving the individual spatial patch
tokens: representations that discard them show substantially weaker
encoding, whereas the CLS + all patches concatenated representation
provides the strongest correspondence for both transformer models. We
therefore use the CLS + all patches concatenated representation for
CLIP-ViT-B/32 and DINOv2-ViT-B/14 in the following model-level
comparison, which evaluates all models using their strongest available
representation under the same encoding framework.

\begin{figure}[pos=htbp]
\centering
\includegraphics[width=\linewidth]
    {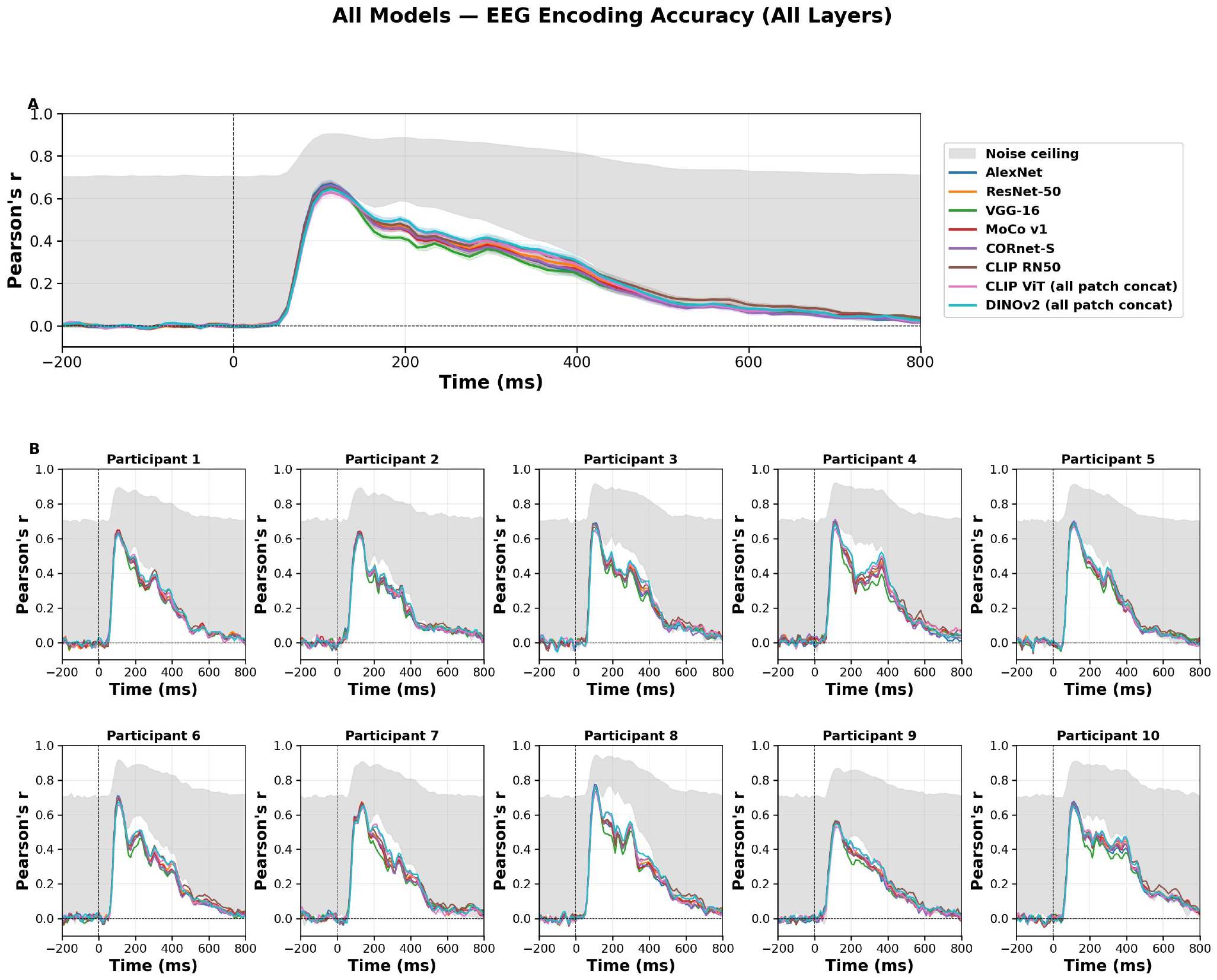}
\caption{Temporal EEG encoding profiles for all base models and transformer token variants using all layer concatenated representations. Panel A: group average across 10 participants; Panel B: individual participant profiles.}
\label{fig:all_models}
\end{figure}

\begin{figure}[pos=htbp]
\centering
\includegraphics[width=0.7\linewidth]
    {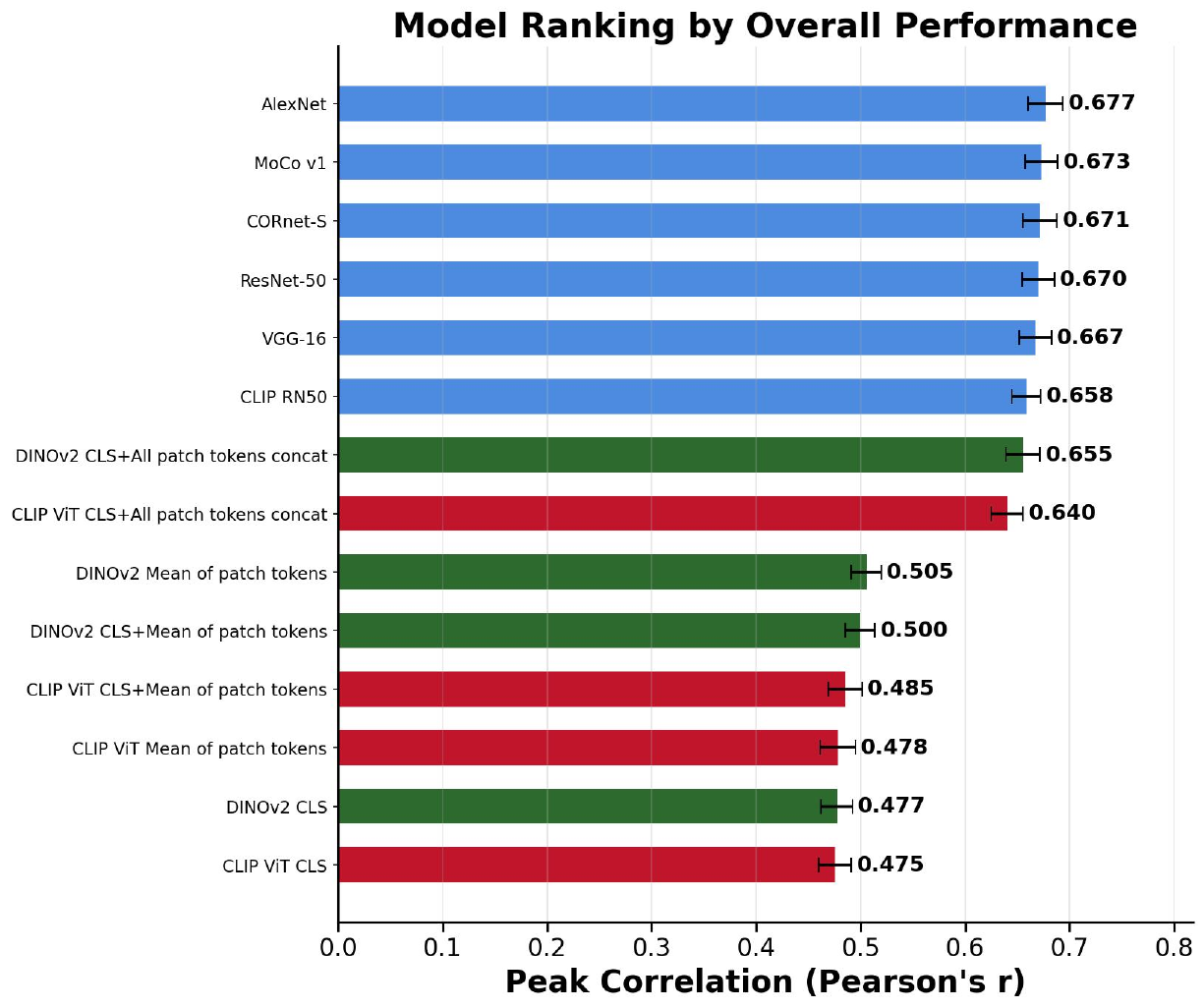}
\caption{Peak EEG encoding accuracy (Pearson r, mean ± SEM over 10 participants) for all base models and transformer token variants, sorted in descending order.}
\label{fig:ranking_bar}
\end{figure}

We next examine encoding performance across the complete set of model
representations, using features from all network depths. For the
transformer models, the CLS + all patches concatenated representation is
used where specified. Fig.~\ref{fig:all_models} shows the temporal
encoding profiles obtained from these representations, while
Table~\ref{tab:ranking} and Fig.~\ref{fig:ranking_bar} summarize the
corresponding peak encoding accuracies.

\FloatBarrier

Across the eight models, the CNN and CNN-backbone models achieve
the highest peak encoding accuracies, with AlexNet achieving
$r{=}0.677 \pm 0.052$, MoCo-v1 $r{=}0.673 \pm 0.050$, CORnet-S
$r{=}0.672 \pm 0.052$, ResNet-50 $r{=}0.670 \pm 0.049$, VGG-16
$r{=}0.667 \pm 0.049$, and CLIP-RN50 $r{=}0.659 \pm 0.043$. Using the
CLS + all patches concatenated representation, DINOv2 achieves
$r{=}0.656 \pm 0.051$, while CLIP-ViT achieves $r{=}0.640 \pm 0.048$. The
resulting ranking differs from the layer-wise comparison in
Sections~\ref{results}A and~\ref{results}B because this analysis combines representations
across all depths. Consequently, strong encoding from early and
intermediate representations contributes substantially to the overall
peak accuracy and can mask the differences observed at deeper layers.

\begin{table}[t]
\caption{Peak EEG encoding accuracy (Pearson $r$) across all models and their respective token variants.}
\label{tab:ranking}
\begin{tabular*}{\tblwidth}{@{}LLL@{}}
\toprule
\textbf{Model} & \textbf{Mean $r$} & \textbf{Peak time (ms)} \\
\midrule
AlexNet                              & $0.677 \pm 0.052$ & $114.14$ \\
MoCo-v1                               & $0.673 \pm 0.050$ & $113.13$ \\
CORnet-S                              & $0.672 \pm 0.052$ & $114.14$ \\
ResNet-50                            & $0.670 \pm 0.049$ & $112.12$ \\
VGG-16                                & $0.667 \pm 0.049$ & $111.11$ \\
CLIP-RN50                             & $0.659 \pm 0.043$ & $114.14$ \\
DINOv2 (CLS + all patches)            & $0.656 \pm 0.051$ & $117.17$ \\
CLIP-ViT (CLS + all patches)          & $0.640 \pm 0.048$ & $114.14$ \\
DINOv2 (mean pooled patches)          & $0.505 \pm 0.045$ & $116.16$ \\
DINOv2 (CLS + mean pooled patches)    & $0.500 \pm 0.044$ & $118.18$ \\
CLIP-ViT (CLS + mean pooled patches)  & $0.485 \pm 0.051$ & $114.14$ \\
CLIP-ViT (mean pooled patches)        & $0.478 \pm 0.052$ & $115.15$ \\
DINOv2 (CLS only)                     & $0.477 \pm 0.047$ & $117.17$ \\
CLIP-ViT (CLS only)                   & $0.476 \pm 0.049$ & $117.17$ \\
\bottomrule
\end{tabular*}
\end{table}

To examine the effect of training objective while keeping the
architecture fixed, we compare MoCo-v1 with ResNet-50, since both use a
ResNet-50 backbone but differ in their training objectives. Their
temporal encoding profiles are highly similar, and their peak encoding
accuracies are also nearly identical, with MoCo-v1 achieving
$r{=}0.673 \pm 0.050$ and ResNet-50 achieving $r{=}0.670 \pm 0.049$. This
close correspondence suggests that, at the level of encoding performance
across the complete set of model representations, changing from
supervised classification to the contrastive objective used by MoCo-v1
does not substantially alter the EEG encoding pattern, both models are unable to capture the semantic representations.

We next compare CLIP-RN50 and CLIP-ViT-B/32, which share the same
contrastive language-image training objective but differ in
architecture. With the standard transformer representation, CLIP-ViT
achieves a lower peak encoding accuracy of $r{=}0.476 \pm 0.049$
compared with $r{=}0.659 \pm 0.043$ for CLIP-RN50. However, despite its
lower peak correlation, the CLIP-ViT temporal profile shows stronger
correspondence during the later portion of the post-stimulus response,
including the period associated with higher level and semantic
processing. The subsequent token representation analysis provides an
explanation for this difference. When the individual patch tokens are
retained using the CLS + all patches concatenated representation,
CLIP-ViT reaches a substantially higher peak encoding accuracy of
$r{=}0.640 \pm 0.048$, approaching the $r{=}0.659 \pm 0.043$ achieved by
CLIP-RN50, while retaining the stronger late temporal encoding profile.
Thus, the transformer advantage is not simply reflected in its maximum
correlation; rather, it is reflected in its ability to maintain
brain-predictive information into later stages of the neural response,
which becomes more apparent when the patch representations are
retained.
\FloatBarrier

\subsection{Spatial Topography of Encoding Accuracy Across Model Categories}

The stage wise and model level analyses above average encoding across all 17 EEG channels. To determine whether this averaging conceals architecture or objective specific spatial patterns, we visualize the full channel × time correlation surface averaged across all 10 participants.

\begin{figure}[pos=htbp]
\centering
\includegraphics[width=0.7\linewidth]
    {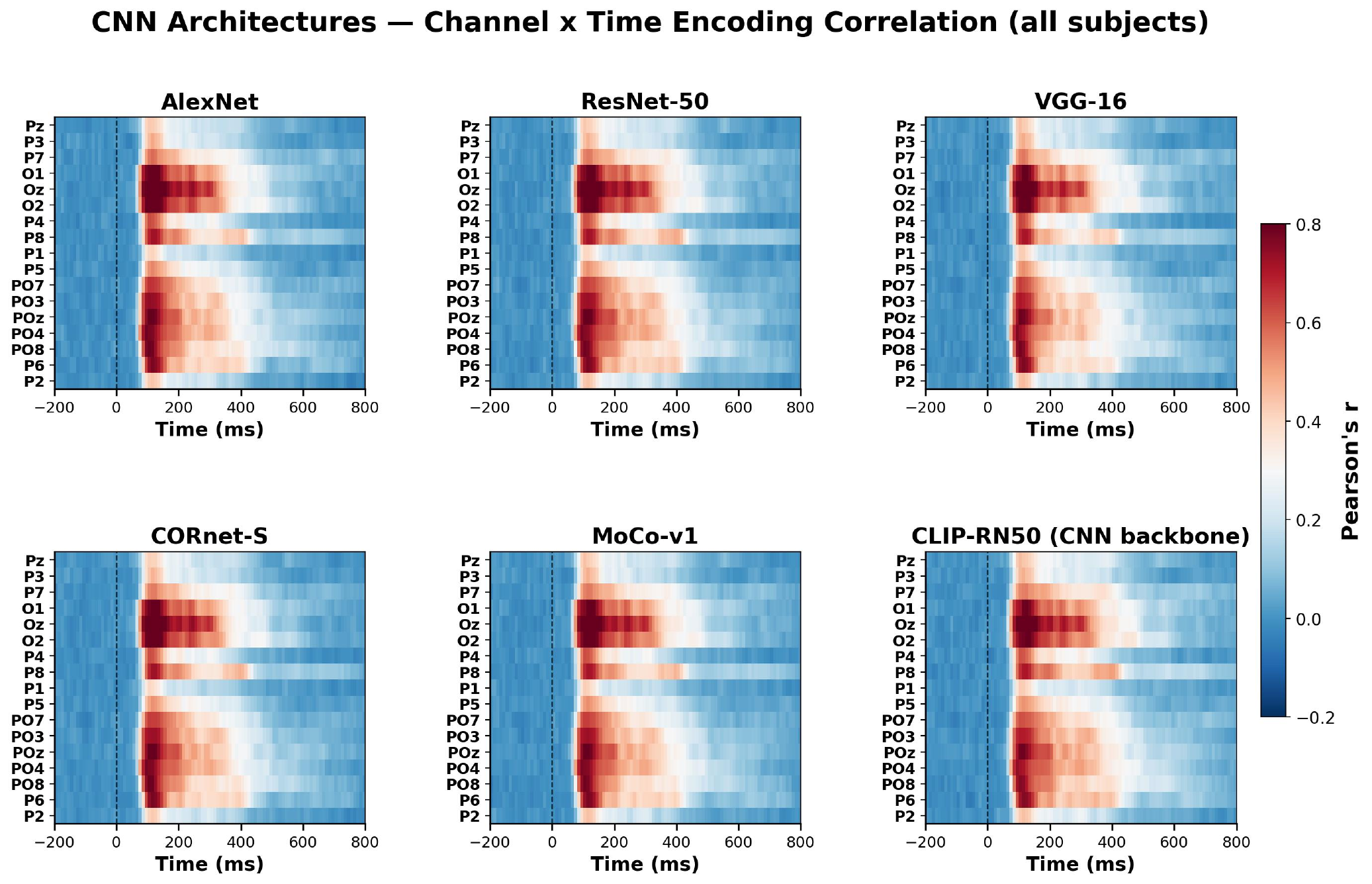}
\caption{Channel × time EEG encoding correlation for CNN and CNN-backbone models (all subjects)}
\label{fig:heatmap_cnn}
\end{figure}

\textbf{Supervised and self-supervised CNNs.} Fig.~\ref{fig:heatmap_cnn}
shows AlexNet, ResNet-50, VGG-16, CLIPRN50, CORnet-S, and MoCo-v1. All models
produce a visually near identical topography: encoding accuracy peaks
sharply in the occipital channels (O1, Oz, O2),
falls off moderately in the parieto occipital channels (PO7, PO3, POz, PO4,
PO8), and remains weak throughout the parietal channels (Pz, P1--P8). The
onset is sharp at 60--70 ms in every model. This topography is unchanged
between classification trained networks (AlexNet, VGG-16, ResNet-50,
CORnet-S) and the contrastively trained MoCo-v1, indicating that this
spatial signature is a property of the convolutional architecture rather
than the supervised/self-supervised distinction.

\textbf{CNN backbone with a non classification objective.} CLIP-RN50, shown
in the sixth panel of Fig.~\ref{fig:heatmap_cnn}, reproduces the same
occipital dominant topography as the classification trained CNNs despite
never being trained to classify images. Holding architecture fixed while
changing the training objective leaves the spatial pattern of brain
alignment unchanged, reinforcing that this topography tracks architecture,
not objective.

\FloatBarrier

\begin{figure}[pos=htbp]
\centering
\includegraphics[width=0.8\linewidth]
    {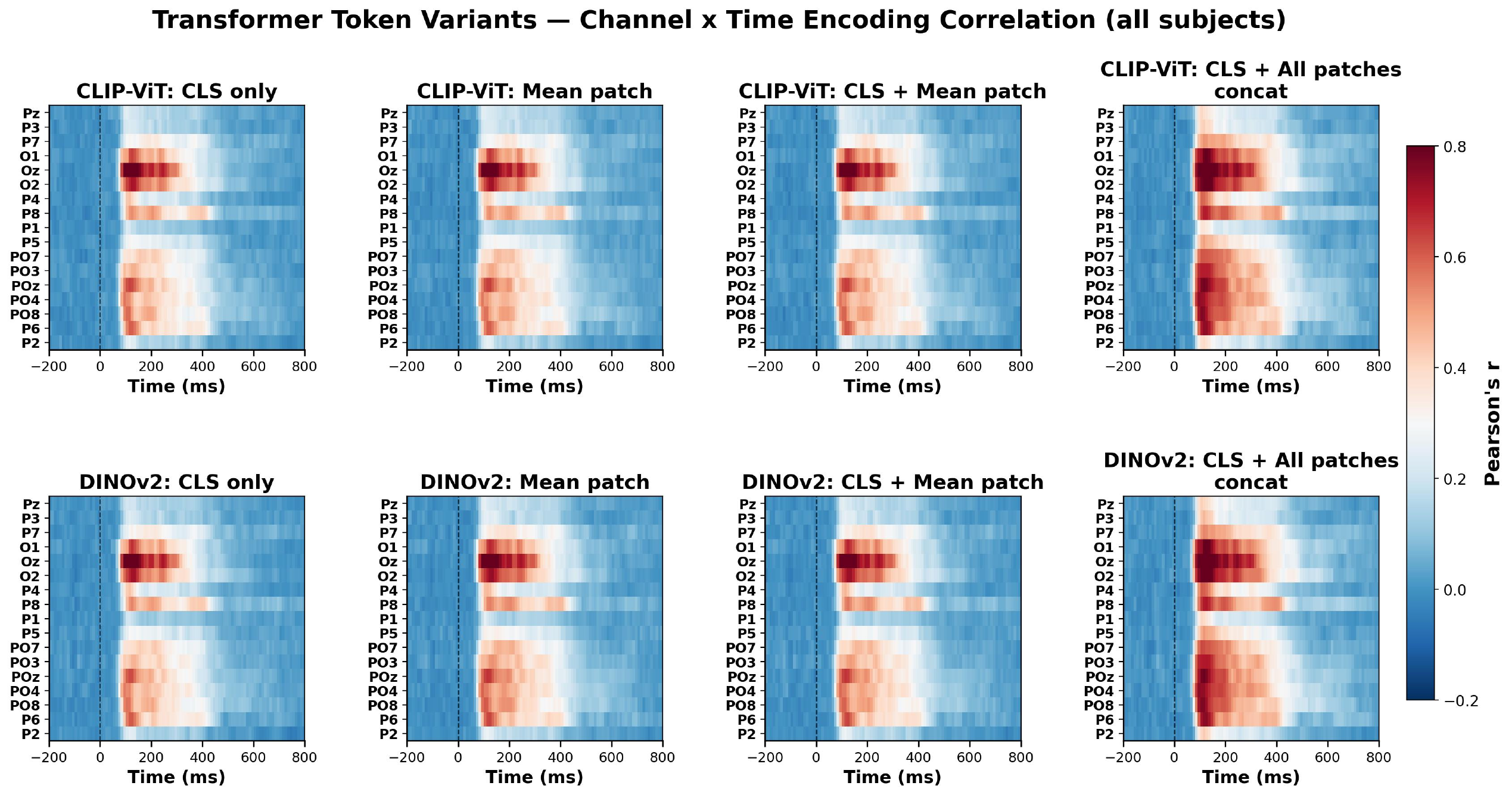}
\caption{Channel × time EEG encoding correlation for four token representations of CLIP-ViT-B/32 and DINOv2-ViT-B/14 (all subjects).}
\label{fig:heatmap_tokens}
\end{figure}

\textbf{Transformer token variants.} Fig.~\ref{fig:heatmap_tokens}
shows all four token strategies for both CLIP-ViT-B/32 and DINOv2-ViT-B/14. Every
variant preserves the same occipital dominant spatial topography; token
aggregation choice does not change where the signal is strongest.
What changes is the strength and duration of the response: the CLS+all patches concatenated
variant shows the deepest and longest sustained correlation (extending to
roughly 400--500 ms in the parieto-occipital channels), while the pooled
variants (CLS only, mean pooled patches) show a visibly weaker, more
short lived response. This provides direct spatial temporal confirmation
of the token aggregation result in Section~\ref{results}: retaining
individual patch tokens, not a pooled summary, is what sustains encoding
accuracy.

\FloatBarrier

Taken together, these four categories show that the occipital dominant
spatial topography is present in visual encoding task in all models and variants tested, and does not by
itself distinguish CNNs from transformers or classification objectives from
contrastive ones. The architecture and objective dependent effects reported
throughout this study the semantic stage collapse of CNNs, the sustained
accuracy of transformers, and the dependence of that sustained accuracy on
patch token retention are differences in the temporal extent and strength
of the encoded signal within this shared spatial map, not differences in
which channels respond.

\subsection{Discussion}

The findings show a distinct difference in the way CNNs and vision
transformers encode visual information as a function of network depth
and time. In the case of CNNs, the strongest EEG correlation is found
in the first layers, which is in line with the initial feedforward
phase of low-level visual processing, although this correlation
gradually decreases at deeper layers, especially in the later part of
the post-stimulus response. This trend is observed regardless of
whether the CNNs were trained in a supervised, self-supervised, or
contrastive manner, and it is clearly visible in the layer $\times$
time heatmaps of Fig.~\ref{fig:panel_cnn_heatmaps}, where the highest
correlation values are located in the early layers right after the
stimulus has been presented, while the deepest layers stay
comparatively dark over the entire period. Vision transformers, on the
other hand, display a different pattern: rather than showing a simple
decline with increasing depth, their deeper blocks maintain a
relatively strong correlation with the EEG response right up to the
later part of the time window, as can be seen in the block $\times$
time heatmaps of Fig.~\ref{fig:panel_dinov2_heatmaps} and
Fig.~\ref{fig:panel_clip_vit_heatmaps}. It should be noted, though,
that the earliest transformer blocks do not reach the level of
correlation found in the earliest CNN layers, so the difference between
the two types of networks is not that one of them uniformly performs
better than the other; it is that CNNs and transformers distribute
their correspondence with the brain in different ways across depth and
time.

It is important to note that this advantage of the transformer relies
strongly on how the internal token representation is kept. When only
the CLS token or a pooled summary of the patch tokens is used, both
CLIP-ViT-B/32 and DINOv2-ViT-B/14 exhibit considerably lower peak
correlations (approximately $r=0.48$ to $0.51$) than in the case where
the full set of individual patch tokens is retained together with the
CLS token ($r=0.640$ for CLIP-ViT and $r=0.656$ for DINOv2). The same
advantage is evident in the block $\times$ time heatmaps and in the
spatial topography analysis, since the representation that preserves
the patches maintains a more robust and prolonged response than the
ones based on pooling. Therefore, the transformer advantage is not an
inherent feature of the architecture by itself, but rather arises
specifically when the distributed patch-level information is preserved
rather than compressed.

The controlled comparisons allow the contributions of architecture and
training objective to be separated. MoCo-v1 and ResNet-50 have the
same convolutional architecture but different training objectives, and
their whole model peak accuracies are very similar ($r=0.673$ and
$r=0.670$), and their temporal profiles are also highly similar, which
together show that altering the training objective alone, while
keeping the CNN architecture fixed, does not make a significant
difference to the encoding pattern. On the other hand, CLIP-RN50 and
CLIP-ViT-B/32 have the same contrastive language-image objective but
differ in architecture. With its default representation, CLIP-ViT
achieves a considerably lower peak accuracy ($r=0.476$) than CLIP-RN50
($r=0.659$), although its temporal profile already displays a stronger
correspondence later in the post-stimulus response. When the individual
patch tokens are retained, CLIP-ViT's peak accuracy increases to
$r=0.640$, reaching a level close to that of CLIP-RN50 while still
maintaining its stronger late temporal profile. This suggests that the
transformer's ability for sustained late correspondence is dependent on
its architecture, but in order to fully show off that advantage in peak
accuracy it is also necessary to preserve the patch level
representation structure.

The analysis of spatial topography offers a complementary perspective.
Whether or not the models and token variants considered differ in
architecture, training objective, or token aggregation strategy, they
all produce the same occipital-dominant spatial pattern, the accuracy
of the encoding being concentrated in the occipital and
parieto-occipital channels. This shows that the differences noted in
this study are not due to the various models activating different
brain areas, but instead result from variations in the strength and
duration of the encoded signal within a common spatial map. The
patch-preserving transformer representation maintains this response for
a long period, approximately 400 to 500 ms, whereas the
pooled representations and the deeper CNN layers exhibit a relatively
weaker and shorter-lasting response.

Together, these results show that both CNNs and transformers are about
equally aligned with the brain's earliest visual response, but they
begin to differ as processing goes deeper and as time moves beyond the
first feedforward pass. In the deepest layers, the alignment between
CNNs and the EEG response fades substantially, whereas
transformers, especially when their patch-level representations are
kept, maintain a stronger correspondence for a longer time. This
benefit is due to a combined effect of both architecture and
representation structure, not to either of these factors by
themselves.

\section{Conclusion}

The study presents a systematic approach for determining when and at what network depth visual models most closely match human brain activity. It finds that both CNNs and transformers show similar levels of alignment in their early layers but become different at deeper layers, transformers retaining a stronger correspondence during the later part of the post-stimulus response, a phase linked to higher level visual areas such as V4 and the inferotemporal (IT) cortex. This benefit is dependent on the transformer token representation employed: by keeping all the patch tokens the overall accuracy of the encoding is considerably improved, which in turn allows for both strong peak correlations and a longer-lasting response. The results also indicate that the classification bottleneck present in conventional CNNs may be responsible for the loss of brain predictive information at deeper layers, since the representations become more and more compressed as they approach fixed categorical outputs. These findings form the basis for creating vision models that are inspired by and aligned with the brain. Instead of preferring one architecture, both CNNs and transformers, as well as possibly hybrid architectures, can offer better representations throughout the visual hierarchy.

\printcredits

\bibliographystyle{cas-model2-names}

\bibliography{references}
\end{document}